\documentclass{article}

\usepackage{microtype}
\usepackage{graphicx}
\usepackage{subfigure}
\usepackage{booktabs} % for professional tables

\usepackage{hyperref}

\usepackage[accepted]{icml2025}

\usepackage{amsmath}
\usepackage{amssymb}
\usepackage{mathtools}
\usepackage{amsthm}
\usepackage{array}
\usepackage{caption}
\usepackage{algorithm}
\usepackage{algorithmic}
\usepackage{graphicx}
\usepackage{booktabs}
\usepackage{xcolor}
\usepackage{multirow}
\usepackage{amsfonts}
\usepackage{xcolor}
\usepackage{colortbl}
\usepackage{colortbl}
\usepackage{makecell}
\usepackage{float}

\definecolor{darkgreen}{rgb}{0.0, 0.5, 0.0}
\definecolor{text_red}{RGB}{220,20,60}

\newcommand{\R}[1]{\textcolor{text_red}{$ ^{#1}$}}
\newcommand{\G}[1]{\textcolor{darkgreen}{$\kern 0.15em ^{#1}$}}
\newcommand{\GG}[2]{\textbf{#1}\textcolor{darkgreen}{$\kern 0.15em ^{#2}$}}
\usepackage[capitalize,noabbrev]{cleveref}

\theoremstyle{plain}

\theoremstyle{definition}

\theoremstyle{remark}

\usepackage[textsize=tiny]{todonotes}

\begin{document}

\twocolumn[
\icmltitle{Poison as Cure: Visual Noise for Mitigating Object Hallucinations in LVMs}
 % Poison as Cure: Visual Noise to Mitigate Object Hallucinations in LVMs

\begin{icmlauthorlist}

\icmlauthor{Kejia Zhang}{1}
\icmlauthor{Keda Tao}{2}
\icmlauthor{Jiasheng Tang}{3,4}
\icmlauthor{Huan Wang}{2}

\end{icmlauthorlist}

\icmlaffiliation{1}{The Department of Artificial Intelligence, Xiamen University}
\icmlaffiliation{2}{School of Engineering, Westlake University}
\icmlaffiliation{3}{DAMO Academy, Alibaba Group}
\icmlaffiliation{4}{Hupan Laboratory}

\begin{center}
\texttt{\url{https://kejiazhang-robust.github.io/poison-cure-lvm}}
\end{center}

\vspace{-0.8em}

\icmlcorrespondingauthor{Huan Wang}{wanghuan.westlake.edu.cn}
\icmlkeywords{Machine Learning, ICML}
% \vspace{-2em}
\vskip 0.25in

{\renewcommand\twocolumn[1][]{#1}
\begin{center}
\centering
\renewcommand{\arraystretch}{0.05} %
\vspace{-8mm}
\begin{tabular}{c}
\hspace{-0.40cm}
% , height = 0.35\linewidth
\includegraphics[width = 0.55\linewidth]{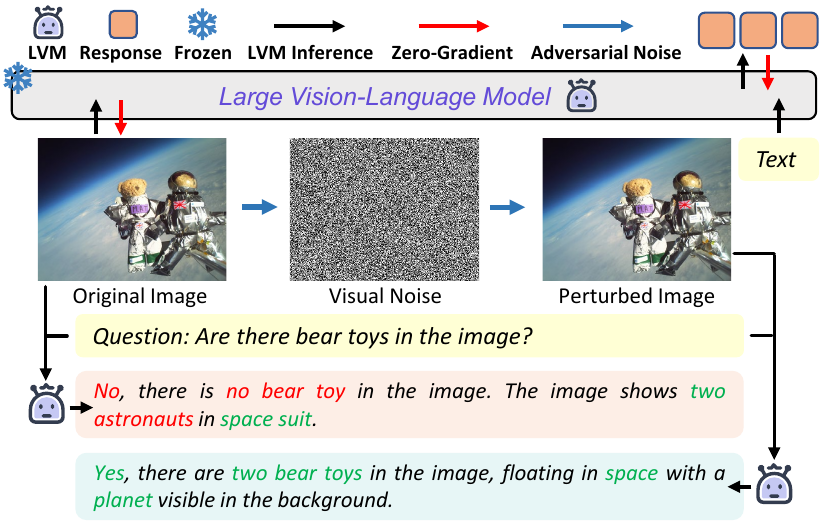} 
\hspace{0.0cm}
% ,height = 0.3\linewidth
\includegraphics[width = 0.45\linewidth]{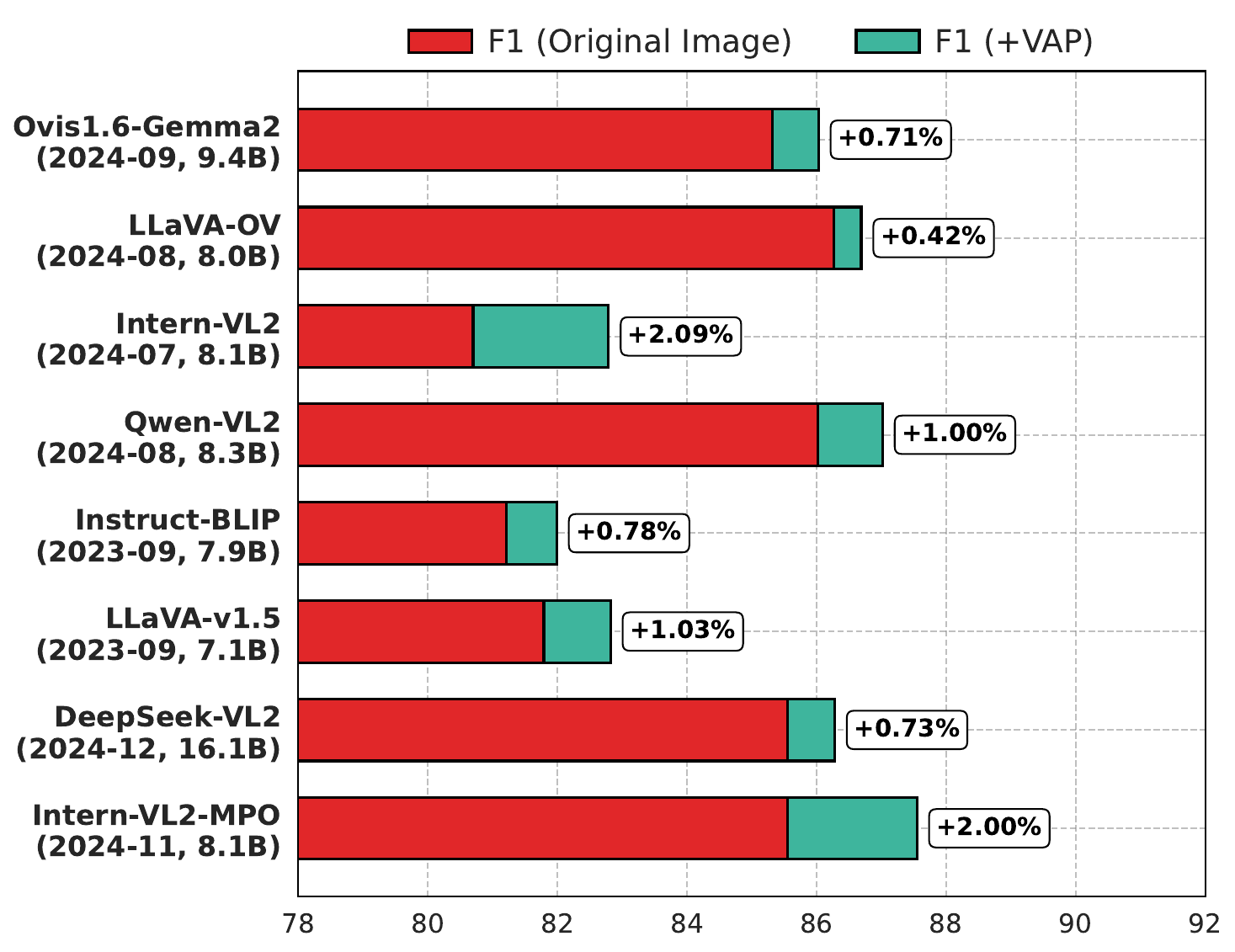}
\end{tabular}
\captionof{figure}{
We introduce \textit{VAP}~(visual adversarial perturbation), a novel approach that strategically injects beneficial visual noise to mitigate object hallucination in LVMs without altering the complex base model. Our method consistently improves performance across 8 state-of-the-art LVMs under the POPE hallucination evaluation setting~\cite{pope}.
}
\label{fig:teaser}
\end{center}
}
]
\printAffiliationsAndNotice{}
\begin{abstract}
    Large vision-language models (LVMs) extend large language models (LLMs) with visual perception capabilities, enabling them to process and interpret visual information. A major challenge compromising their reliability is object hallucination that LVMs may generate plausible but factually inaccurate information. We propose a novel \textit{visual adversarial perturbation (VAP)} method to mitigate this hallucination issue. VAP alleviates LVM hallucination by applying strategically optimized visual noise without altering the base model. 
    Our approach formulates hallucination suppression as an optimization problem, leveraging adversarial strategies to generate beneficial visual perturbations that enhance the model's factual grounding and reduce parametric knowledge bias.
    Extensive experimental results demonstrate that our method consistently reduces object hallucinations across 8 state-of-the-art LVMs, validating its efficacy across diverse evaluations.
\end{abstract}

\section{Introduction}
\label{Intro}
Large vision-language models (LVMs) integrate visual and textual information, providing transformative capabilities for addressing complex cross-modal understanding challenges~\cite{Thrush_2022_CVPR, chen2025sharegpt4v, Kuckreja_2024_CVPR}. 
Despite their remarkable advancements, LVMs often generate plausible yet factually inaccurate outputs, eliciting harmful content such as misinformation or biased representations~\cite{pope, menon2024task}. Addressing these limitations is critical to enhancing the reliability and applicability of LVMs in real-world scenarios.

Prior research indicates that hallucinations in LVMs arise from the interaction between biased parametric knowledge and real-world data distributions~\cite{bai2024hallucination, Guan_2024_CVPR, deletanglanguage}. This phenomenon is driven by two primary mechanisms. First, the long-tail distribution of training data induces systematic biases in parametric knowledge, resulting in spurious correlations and factual inconsistencies~\cite{pope, liu2023mitigating}. Second, the extensive parameter spaces of large language models (LLMs) within LVMs amplify these biases, particularly given the LLMs' predominant role in the inference pipeline~\cite{laurenccon2024matters, liu2025paying}. This LLM dominance potentially suppresses critical visual signals, increasing hallucination frequency~\cite{rohrbach2018object, Leng_2024_CVPR}. Consequently, the embedded biased parametric knowledge substantially compromises LVMs' capacity to accurately process real-world data.

Existing solutions to mitigate this challenge have primarily followed two strategies: fine-tuning~\cite{liu2023mitigating, Yu_2024_CVPR, anonymous2024perturbollava} and decoder process optimization~\cite{Huang_2024_CVPR, liu2025paying, chenhalc}. These approaches represent model-centric interventions, which modify LVMs' internal mechanisms through either parametric updates via fine-tuning or algorithmic refinements in the decoding process~\cite{liu2024survey}. 
These approaches have demonstrated substantial success in reducing hallucinations in LVMs, establishing important foundations for improving LVM reliability.

Unlike prior model-centric approaches, we introduce a paradigm shift in hallucination mitigation that leverages the intrinsic mechanisms of hallucinations to suppress them.
This perspective stems from a crucial observation that while hallucinations arise from biased parametric knowledge, they manifest specifically during the processing of real-world visual inputs~\cite{gunjal2024detecting, bai2024hallucination}. This understanding reveals an elegant solution: strategically crafted perturbation to visual inputs can redirect LVMs' decision-making processes away from parametric biases without altering the original model's architecture or mechanisms. 

This insight motivates our visual adversarial perturbation strategy, where adversarial optimization through zero-gradient techniques introduces beneficial visual noise to the original image. This noise guides the model to ground its responses in actual visual content rather than relying on parametric knowledge biases. The power of this approach lies in its exploitation of visual inputs as concrete factual anchors, fundamentally different from language prompts that often reinforce existing parametric biases~\cite{Shtedritski_2023_ICCV, Xiao_2024_CVPR}. Notably, our method functions in a fully black-box manner requiring no access or modification to the LVM, making it a practical and efficient solution.

Building on this foundation, we propose visual adversarial perturbation (VAP), a novel technique designed to mitigates hallucinations by applying beneficial adversarial perturbations to visual inputs~(as shown in \Cref{fig:teaser} (left)). 
Adversarial perturbations, traditionally considered as ``poison'' due to their initial disruption of model decisions, are reformulated to specifically align model responses with visual content and mitigate parametric knowledge bias. 
By adversarially optimizing visual noise, VAP refines LVM decision-making in a data-centric manner, transforming perturbations from a factor of degradation into a corrective ``cure'' that effectively mitigates object hallucinations.
% VAP leverages adversarially optimized visual noise to refine LVM decision-making through a data-centric approach. This optimization transforms perturbations from a disruptive factor into a ``cure'' that actively corrects hallucinations, effectively mitigating object hallucinations.
% VAP strategically optimizes visual noise through adversarial strategies to influence LVM decision-making, effectively reducing object hallucinations via a data-driven approach.
% This strategic influence reshapes perturbations from a source of degradation into a ``cure'', actively suppressing hallucinations and improving model reliability.

We evaluate the effectiveness of VAP using complementary hallucination assessment frameworks: POPE~\cite{pope} and BEAF~\cite{ye2024beaf} for closed VQA evaluation, and CHAIR~\cite{rohrbach2018object} for open-ended generation tasks. Our extensive experiments across 8 state-of-the-art (SOTA) LVMs demonstrate that VAP consistently mitigates hallucinations across diverse evaluation settings.

Overall, our contributions are structured as follows:
\begin{itemize}
    \item We propose visual adversarial perturbation, a novel method for mitigating object hallucinations in LVMs through beneficial adversarial perturbations applied to visual inputs, without modifying intricate LVMs.
    \item We formulate object hallucination mitigation as an adversarial visual noise optimization. By refining adversarial strategies, beneficial visual noise is generated through zero-gradient optimization to influence model decision-making and alleviate hallucinations.
    \item Extensive experiments across multiple evaluation settings, including text-axis, text- and vision-axes, and open-ended image caption generation, validate the efficacy of our method in reducing hallucinations.
\end{itemize}
\section{Related Work}

\subsection{Large-Vision Language Models}
In recent years, the field has witnessed significant advancements in large vision-language models (LVMs). Numerous LVMs have been developed to tackle real-world multimodal challenges such as image captioning and visual question answering~\cite{xu2024lvlm, wang2024visionllm}. 
LVMs typically operate through a structured pipeline comprising a visual encoder, a cross-modal connector, and a large language model (LLM), facilitating seamless interaction between visual and linguistic features.
State-of-the-art approaches leverage extensive datasets and employ a two-stage training paradigm: pretraining on diverse multimodal corpora~\cite{clip, lion}, followed by fine-tuning with task-specific instructions~\cite{llava, instruct_tuning_paper}. This methodology enables LVMs to interpret and respond to complex multimodal inputs with remarkable efficacy~\cite{llava_ov, instruct_blip}.

\subsection{Hallucination in LVMs}
In the realm of LVMs, hallucination refers to the generation of textual responses that deviate from or contradict the actual visual content, leading to factual inaccuracies or biased information~\cite{pope, Biten_2022_WACV, bai2024hallucination}. These hallucinations primarily arise from intrinsic limitations of LVMs, specifically: (1) the long-tail distribution of training data, which introduces systematic biases into the model's parametric knowledge~\cite{zhouanalyzing, Yu_2024_CVPR}; and (2) the vast parameter space of LLMs, which dominate the inference process and exacerbate these biases~\cite{liu2024survey, liu2025paying}. Due to the fundamental role of objects in computer vision and multimodal research, current evaluation frameworks primarily concentrate on object hallucination~\cite{rohrbach2018object,zhouanalyzing}.

Prior work has explored two primary model-centric strategies to mitigate object hallucinations in LVMs: fine-tuning and decoding strategies. These interventions target the underlying parametric knowledge bias that leads to hallucinations. 
Fine-tuning approaches like REVERIE~\cite{kim2025exploiting} and HalluciDoctor~\cite{Yu_2024_CVPR} update the parametric knowledge of LVMs through comprehensive instruction data to suppress hallucinations. 
Meanwhile, decoding-based methods such as PDM~\cite{favero2024multi} and OPERA~\cite{Huang_2024_CVPR} mitigate hallucinations by intervening in the model's decoding process. In contrast to these model-centric strategies, we approach the challenge from a data-centric perspective, proposing a novel adversarial visual perturbation technique that directly mitigates object hallucinations through visual perturbations.

% 1. CVPR-24 OPERA: 幻觉是由于模型解码时过分关注语言先验。~\cite{Huang_2024_CVPR} （decoding）
% 2. ICLR-24 LRV-Instruction: 过分关注语言先验 ~\cite{liu2023mitigating}
% 导致生成与instruct text更有肯呢个结合的单词而不管图像内容。 （微调）
% 3. CVPR-24 HalluciDoctor: 机器生成数据是长尾分布偏差会导致幻觉，导致错误的推断不存在的物体的出现（还是语言先验）。 设计||设计反事实指令抑制幻觉。（微调）~\cite{Yu_2024_CVPR}
% 4. ECCV-24 Paying more attention: LVMs受到文本惯性“过度依赖LLMs,有些问题仅受上下文影响”,解码过程调整注意力中减去LLMs偏差。 (decoding) ~\cite{liu2025paying}

% 在本文中 我们专注于利用adversarial attck幻觉object hallucination, and leave fine-grained object hallucinations such as the number, attributes, and positions of the object for future work.

% Improving the decoding process or fine-tuning LVMs often involves substantial complexity and computational cost. To address this, we propose a novel approach to mitigate hallucinations without altering the model architecture or requiring fine-tuning. Our method introduces adversarial perturbations to guide the model’s outputs more effectively, offering an efficient alternative to traditional optimization techniques.

\begin{figure*}[t]
        \begin{center}
        \includegraphics[width=\linewidth]{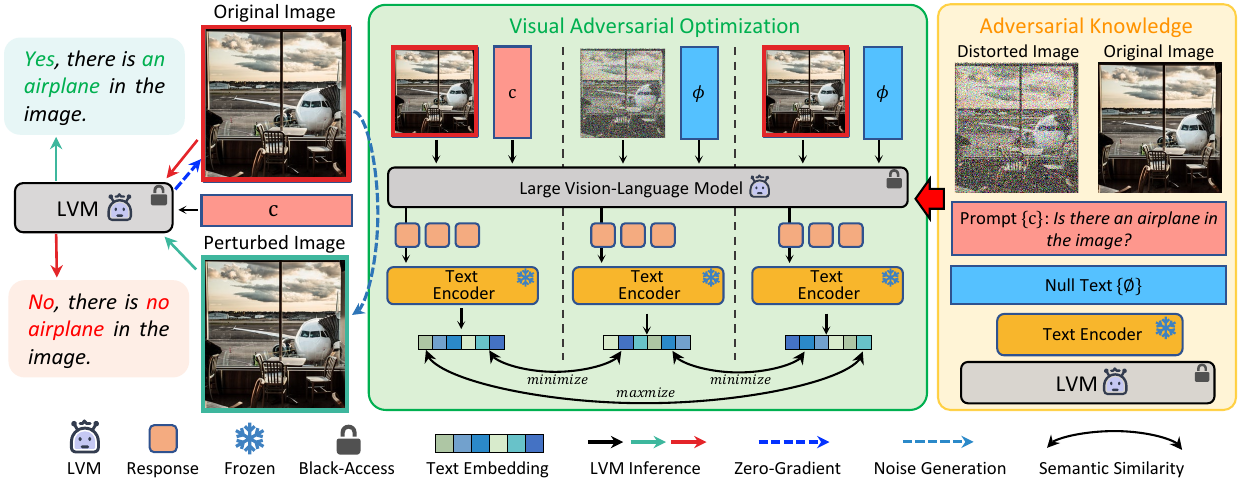}
        \end{center}
        \caption{\textbf{Detailed Overview of our proposed method.} The VAP method generates beneficial visual noise by leveraging adversarial knowledge through the optimization of three strategies: (1) maximizing the semantic alignment between the LVM's response and the visual content to preserve the semantic consistency of the image, (2) minimizing the response similarity between the original and distorted visual content through noise-induced uncertainty, and (3) mitigating parametric knowledge bias by minimizing the similarity of representations between original and distorted inputs. Strategies (2) and (3) jointly mitigate parametric knowledge bias. The optimized visual noise effectively mitigates object hallucinations.}
        \label{VAP_framework}
\end{figure*}

\section{Methodology}
We propose visual adversarial perturbation (VAP) to mitigate object hallucination in LVMs. VAP formulates an adversarial strategy to align the LVM responses with visual content while reducing the impact of parametric knowledge bias~(\Cref{sec:strategy}). These objectives guide the adversarial optimization process, which generates beneficial visual noise to improve model performance~(\Cref{sec:optimization}). An overview of our framework is shown in \Cref{VAP_framework}.

\subsection{Preliminaries}
\textbf{Notations}
Let $f_\theta$ denote LVM, where $x$ represents the input image, $c$ is the query prompt, and $w$ is the model's generated response, such that $w = f_\theta(x, c)$. We define $g_{\psi}$ as the CLIP text encoder converting textual data into semantically meaningful embeddings. For adversarial perturbation, we denote $\delta$ as the perturbation vector and $\mathcal{L}_{S}$ as the surrogate adversarial loss guided by strategy set $S=[s_1,\cdots,s_n]$. The perturbed image is defined as $\hat{x}=x+\delta$, $\epsilon$ is the magnitude of perturbation, and ${\Omega}$ represents the adversarial knowledge utilized during the adversarial optimization process.

\textbf{Adversarial Perturbation}
Adversarial perturbation against LVMs typically involves adding imperceptible visual noise to influence model decisions~\cite{attack_neurips_2023, Cui_2024_CVPR}, which can significantly alter the model's output. The optimization of such perturbations can be formulated as:
\begin{equation}
    \delta = \underset{\delta \sim \mathbb{B}_\epsilon(x)}{\arg \max}\  \mathcal{L}_{(S)} (x+\delta, {\Omega}),
    \label{adversarial_loss_perturbation}
\end{equation}
where $\delta$ represents the adversarial perturbation to be optimized, $\mathcal{L}_{(S)}$ represents the adversarial objective function under strategy $S$, and $\Omega$ indicates the available adversarial knowledge. The perturbation is bounded within an $\epsilon$-ball $\mathbb{B}$.
Specifically, the adversarial perturbation is optimized by computing the gradient as follows:
\begin{equation}
    \hat{x} = x + \alpha \nabla_{x} \{ \mathcal{L}_{(S)} (x + \delta, {\Omega}) \},
    \label{gradient}
\end{equation}
where $\alpha$ is the step size, and the gradient $\nabla_{x}$ is computed with respect to the vision input $x$ to update perturbation $\delta$.

\subsection{Adversarial Strategies}
\label{sec:strategy}
Our adversarial goal is formulated as two principal objectives: \textbf{(1)} optimizing the semantic alignment between the LVM's response and the corresponding visual content, and \textbf{(2)} mitigating the influence of parametric knowledge bias. 

\par \noindent \textbf{Alignment LVM Response with Grounding Visual Content}
Hallucinations in LVMs manifest as the generation of semantically plausible responses but diverge from the actual visual content. 
To mitigate this, our proposed methodology promotes enhanced alignment between the model's responses and the actual visual content:
\begin{equation}
    \mathcal{L}_{s_1} = \underset{\delta \sim \mathbb{B}_\epsilon(x)}{\max} \{S(f_\theta(x+\delta,c), f_\theta(x+\delta,\emptyset))\}, \label{eq:ls1}
\end{equation}
where $S(\cdot,\cdot)$ signifies the calculation of semantic correlation between the two generated responses, $f_\theta(x+\delta,c)$ represents the model's output given the perturbed vision input $x+\delta$ with the conditional query prompt $c$, and $f_\theta(x+\delta,\emptyset)$ signifies the visual semantic description when the prompt is replaced with an empty token $\emptyset$. 
This loss term $\mathcal{L}_{s_1}$ quantifies the semantic alignment between conditionally guided responses and the model's autonomous interpretation of visual content, thereby enhancing response consistency with the underlying visual semantics.

Despite the improvements, the alignment between responses and visual content may still be influenced by parametric knowledge bias, particularly an over-reliance on linguistic priors~\cite{anonymous2024perturbollava}. Such bias can distort the model's interpretation of visual information, leading to hallucinatory patterns. As discussed in \Cref{Intro}, LVMs often prioritize linguistically anchored priors over visual signals, thereby exacerbating existing biases. Our alignment strategy addresses this by mitigating both misalignment and bias.

\par \noindent \textbf{Mitigating Parametric Knowledge Bias}
Visual uncertainty~\cite{Guan_2024_CVPR, Leng_2024_CVPR} serves as a critical metric for quantifying parametric knowledge bias. It is quantified by generating a contrastive negative image $\bar{x}$ through the introduction of noise to the original image:
\begin{equation}
p(\bar{x}|x) = \mathcal{N}(\bar{x}; \sqrt{\mu_T}x, (1-\mu_T)\mathbf{I}),
\label{Distorted_T}
\end{equation}
where $\mu_T$ represents the noise scheduling coefficient at timestep $T$, controlling the magnitude of perturbation applied to the original image $x$.

To further mitigate parametric knowledge bias, we introduce a dual-setting approach that reduces the semantic similarity between LVM responses to original and distorted visual inputs under both conditional $c$ (with query prompt) and unconditional $\emptyset$ (without query prompt) configurations.

In the conditional $c$ setting, our approach minimizes the semantic similarity between the perturbed input $x+\delta$ and the contrastive negative image $\bar{x}$:
\begin{equation}
    \mathcal{L}_{s_2} = \underset{\delta \sim \mathbb{B}_\epsilon(x)}{\min} \{S(f_\theta(x+\delta,c), f_\theta(\bar{x},\emptyset))\}, \label{eq:ls2}
\end{equation}
where $f_\theta(\bar{x},\emptyset)$ denotes the LVM's output given the visually uncertain input. $\mathcal{L}_{s_2}$ promotes discriminative responses between prompted and unprompted conditions, thereby reducing dependency on linguistic priors.

In the unconditional $\emptyset$ setting, our methodology minimizes the semantic similarity between responses to the perturbed image $x+\delta$ and its contrastive negative counterpart $\bar{x}$:
\begin{equation}
    \mathcal{L}_{s_3} = \underset{\delta \sim \mathbb{B}_\epsilon(x)}{\min} \{S(f_\theta(x+\delta,\emptyset), f_\theta(\bar{x},\emptyset))\}, \label{eq:ls3}
\end{equation}
where $\mathcal{L}_{s_3}$ alleviates the propensity to hallucinate, further mitigating the dominant influence of linguistic priors.

The loss terms $\mathcal{L}_{s_1}$, $\mathcal{L}_{s_2}$, and $\mathcal{L}_{s_3}$ collectively regulate LVM responses to ensure consistency with visual content while mitigating parametric knowledge bias in LVMs. We formulate our complete optimization objective as a weighted combination of these loss terms:
\begin{equation}
    \mathcal{L}_{S}(x,c,\theta) = \frac{\mathcal{L}{s_1}}{\sigma_1^2} + \frac{\mathcal{L}{s_2}}{\sigma_2^2} + \frac{\mathcal{L}{s_3}}{\sigma_3^2},
\end{equation}
where $\sigma_i^2$ ($i \in \{1,2,3\}$) are balancing coefficients that modulate the contribution of each loss component. This formulation achieves a dual objective: $\mathcal{L}_{s_1}$ ensures strong semantic alignment between model responses and visual content, while $\mathcal{L}_{s_2}$ and $\mathcal{L}_{s_3}$ collectively mitigate parametric knowledge bias through consistent interpretation across visual perturbations.

\subsection{Visual Adversarial Optimization}
\label{sec:optimization}
To optimize our adversarial objectives $\mathcal{L}_S$, we leverage the CLIP text encoder $g_\psi(\cdot)$ as a surrogate model, capitalizing on its superior discriminative capabilities for textual representation~\cite{wu2024llm2clip}. This approach contrasts with the limited semantic separability in LLM representations:
\begin{equation}
    S(\cdot,\cdot) = g_\psi (\cdot)^\top g_\psi (\cdot),
\end{equation}
where $S(\cdot,\cdot)$ measures the similarity of the LVM’s response under different conditions. Then, we compute the numerical loss $\mathcal{L}_S(x, c, \theta)$, which enables the optimization of the perturbation $\delta$. $\delta$ represents a carefully crafted visual perturbation designed to optimize the strategic objective:
\begin{equation}
    \delta = \nabla_x\{\mathcal{L}_{S}(x,c,\theta, \psi)\}.
\end{equation}

The final adversarial perturbation is generated by adding noise to the input image $x$, yielding the visual adversarial perturbed image $\hat{x}$:
\begin{equation}
    \hat{x} = x + \alpha \cdot \delta= x + \alpha \nabla_x\{\mathcal{L}_{S}(x,c,\theta,\psi)\},
\end{equation}
where $\alpha$ denotes the learning rate of adversarial strategies. 
The generated perturbed image $\hat{x}$ exhibits superior optimization characteristics with respect to the objective $\mathcal{L}_{S}$, outperforming the original images $x$ while meticulously preserving the semantic integrity of vision input.

Due to the autoregressive nature of LVMs, direct gradient computation is challenging. To address this, we optimize the similarity-based loss using a gradient-free method~\cite{attack_neurips_2023, nesterov2017random}, which we term zero-gradient optimization. Specifically, we apply the zero-order optimization technique~\cite{chen2017zoo}, which approximates the gradient by evaluating the loss at perturbed inputs and estimating the optimal perturbation direction:
\begin{equation}
    \begin{aligned}
        \nabla_x\{\mathcal{L}_{S}(x,c,\theta)\}\approx & \frac{1}{N\cdot \beta} \sum_{n=1}^N  \{[ \mathcal{L}_{S}(x+\beta\cdot \gamma _n,c,\theta,\psi)\\
        & -\mathcal{L}_{S}(x,c,\theta,\psi)]  \cdot \gamma_n \} ,
    \end{aligned}
\end{equation}
where $\gamma_n$ is sampled from distribution $P(\gamma)$, $\beta$ controls the sampling variance, and $N$ denotes the number of queries. The term $\gamma_n \sim P(\gamma)$ ensures perturbation diversity through the property $E[\gamma^\top \cdot \gamma] = I$. A detailed step-by-step algorithm of VAP is provided in \Cref{sec:appendix_alg}.

% This targeted perturbation effectively suppresses hallucinations by steering LVM responses toward our strategy.
\section{Experiments}

\subsection{Experiment Setup}

% \noindent \textbf{Implementation Details}
% We evaluate our method on eight state-of-the-art LVMs: LLaVA~\cite{llava}, LLaVA-Onevision (OV)\cite{llava_ov}, Instruct-BLIP\cite{instruct_blip}, Intern-VL2~\cite{InternVL}, Intern-VL2-MPO~\cite{InternVL}, Qwen-VL2~\cite{Qwen}, DeepSeek-VL2~\cite{deepseek_vl2}, and Ovis1.6-Gemma2~\cite{ovis}. In our experiments, we select the following parameters: $\alpha=1/255$, $\beta=8/255$, $N=10$, $T=200$, and $\epsilon=2$. Due to the distinct characteristics of each model, we assign different balancing coefficients $\sigma_i$ (where $i \in {1,2,3}$) for each model. The specific configurations and information for each model are provided and analyzed in the \Cref{appendix}.

\noindent \textbf{Implementation Details}
We evaluated our method on 8 state-of-the-art LVMs: LLaVA~\cite{llava}, LLaVA-Onevision~(OV)~\cite{llava_ov}, Instruct-BLIP~\cite{instruct_blip}, Intern-VL2~\cite{InternVL}, Intern-VL2-MPO~\cite{InternVL}, Qwen-VL2~\cite{Qwen}, DeepSeek-VL2~\cite{deepseek_vl2}, and Ovis1.6-Gemma2~\cite{ovis}. In our experiments, we selected the following parameters: $\alpha=1/255$, $\beta=8/255$, $N=10$, $\epsilon=2$. Due to the distinct characteristics of each LVM, we assigned different balancing coefficients $\sigma_i$ (where $i \in {1,2,3}$) and $T$ for each model. Detailed descriptions of these LVMs, along with their specific configurations and comprehensive analyses, are presented in \Cref{appendix}.
% These models are selected to assess the effectiveness of our method in mitigating object hallucinations and represent a diverse range of LVM architectures, each exhibiting distinct characteristics in visual and language processing. This selection ensures a comprehensive comparison across leading models in the field.

\par \noindent \textbf{Evaluation Benchmark}
Our evaluation is divided into two main categories:
\textbf{(1) Closed VQA format for object hallucination evaluation:} Text-axis evaluation POPE~\cite{pope} and vision-/text-axis evaluation BEAF~\cite{ye2024beaf} settings.  
\textbf{(2) Open-ended task evaluation:} Image caption generation CHAIR~\cite{rohrbach2018object} setting. 
Further evaluation details are provided in \Cref{sec:appendix_evaluation}, and comprehensive examples are presented in \Cref{sec:appendix_demo}.

\textbf{1) POPE:} POPE evaluates hallucinations along the text axis by generating VQA pairs through the manipulation of both questions and answers. We randomly selected 500 samples from the MS-COCO dataset and generated 9,000 evaluation triplets using the three sampling strategies described in POPE. Hallucination assessment is performed using Yes/No responses and evaluated with metrics including accuracy, precision, recall, and F1 score.

\textbf{2) BEAF:} BEAF evaluates hallucinations along both the vision and text axes by simultaneously manipulating scene information and questions, enabling a fine-grained hallucination analysis. In addition to Accuracy, Precision, Recall, and F1 score, BEAF incorporates change-aware metrics such as TU, IG, SB\(_p\), SB\(_n\), ID, and F1\(_\text{TUID}\), offering a comprehensive evaluation of object hallucinations. The dataset consists of 26,064 evaluation triplets.

\textbf{3) CHAIR:} CHAIR evaluates hallucination by having the model generate captions and calculating the proportion of objects that appear in the captions but not in the images. Specifically, we randomly selected 1,000 samples from the MS-COCO dataset for evaluation. The assessment uses the following two metrics:
\begin{gather}
    \text{CHAIR}_I = \frac{|{\text{hallucinated objects}}|}{|{\text{total objects mentioned in captions}}|}, \\
    \text{CHAIR}_S = \frac{|{\text{captions with hallucinated objects}}|}{|{\text{total captions generated}}|},
\end{gather}
where $\text{CHAIR}_I$ is calculated at the object level, and $\text{CHAIR}_S$  is calculated at the sentence level.
\begin{table*}[t]
    \begin{center}
    % \caption{Performance comparison under the POPE evaluation setting using popular, random, and adversarial sampling strategies~(\textcolor{red}{explain}). The values in green represent the percentage improvements achieved by our proposed method.}
    \caption{Text-axis evaluation comparison under three evaluation settings of POPE on the validation set of MSCOCO: Random Sampling (selecting absent objects), Popular Sampling (choosing the most frequent missing objects based on dataset-wide occurrence), and Adversarial Sampling (ranking objects by co-occurrence with ground-truth and selecting the most frequent ones). The values in green indicate the percentage improvements achieved by our proposed method.}
    \label{st:SOTA_Comparison}
    \renewcommand{\arraystretch}{1}  % Increase row spacing
    \resizebox{\textwidth}{!}
    {
        \begin{tabular}{l c *{6}{p{1.9cm}}}
        \toprule[1.2pt]
        \multirow{2}{*}{\textbf{LVM}} & 
        \multirow{2}{*}{\textbf{Vision Input}} & 
        \multicolumn{2}{c}{\textbf{Popular}} & 
        \multicolumn{2}{c}{\textbf{Random}} & 
        \multicolumn{2}{c}{\textbf{Adversarial}} \\
        \cmidrule[0.5pt](lr){3-4} \cmidrule[0.5pt](lr){5-6} \cmidrule[0.5pt](lr){7-8}
        & & \textbf{Acc.$\uparrow$} & \textbf{F1$\uparrow$} & 
        \textbf{Acc.$\uparrow$} & \textbf{F1$\uparrow$} & 
        \textbf{Acc.$\uparrow$} & \textbf{F1$\uparrow$} \\ 
        \midrule[0.8pt]

        \multirow{2}{*}{LLaVA-v1.5} &
        \textit{Original} & 85.57 & 86.19 & 88.97 & 89.09 & 79.80 & 81.79 \\
        & \textit{+AVP} & \GG{86.67}{+1.10} & \GG{87.18}{+0.99} & \GG{90.00}{+1.03} & \GG{90.07}{+0.98} & \GG{80.97}{+1.17} & \GG{82.82}{+1.03} \\ 
        \midrule[0.4pt]

        \multirow{2}{*}{Instruct-BLIP} & 
       \textit{Original} & 83.30 & 82.85 & 88.13 & 87.18 & 81.33 & 81.21 \\
        & \textit{+AVP} &  \GG{84.06}{+0.76} & \GG{83.67}{+0.82} & \GG{89.00}{+0.87} & \GG{88.12}{+0.99} & \GG{82.03}{+0.70} & \GG{81.99}{+0.78} \\ 
        \midrule[0.4pt]

        \multirow{2}{*}{Intern-VL2} & 
       \textit{Original} & 84.11 & 81.64 & 85.14 & 82.60 & 82.00 & 80.70 \\
        & \textit{+AVP} & \GG{86.18}{+2.07} & \GG{84.19}{+2.00} & \GG{86.30}{+1.16} & \GG{84.08}{+1.48} & \GG{84.81}{+2.81} & \GG{82.79}{+2.09} \\ 
        \midrule[0.4pt]

        \multirow{2}{*}{Intern-VL2-MPO} &
       \textit{Original} & 87.51 & 86.53 & 88.68 & 87.58 & 86.28 & 85.55 \\
        & \textit{+AVP} & \GG{89.08}{+1.57} & \GG{88.27}{+1.74} & \GG{90.20}{+1.52} & \GG{89.30}{+1.72} & \GG{88.13}{+1.85} & \GG{87.55}{+2.00} \\ 
        \midrule[0.4pt]

        \multirow{2}{*}{DeepSeek-VL2} & 
       \textit{Original} & 86.80 & 85.86 & 88.70 & 87.64 & 86.47 & 85.55 \\
        & \textit{+AVP} & \GG{87.60}{+0.80} & \GG{86.70}{+0.84} & \GG{89.30}{+0.60} & \GG{88.31}{+0.67} & \GG{87.13}{+0.66} & \textbf{86.28}\G{+0.73} \\ 
        \midrule[0.4pt]

        \multirow{2}{*}{Qwen-VL2} & 
       \textit{Original} & 88.13 & 87.68 & 90.60 & 89.99 & 86.27 & 86.02 \\
        & \textit{+AVP} & \GG{89.10}{+0.97} & \GG{88.65}{+0.97} & \GG{91.16}{+0.56} & \GG{90.54}{+0.55} & \GG{87.30}{+1.03} & \GG{87.02}{+1.00} \\ 
        \midrule[0.4pt]

        \multirow{2}{*}{LLaVA-OV} &
       \textit{Original} & 88.30 & 87.33 & 89.53 & 88.51 & 87.17 & 86.27 \\
        & \textit{+AVP}  & \GG{88.93}{+0.63} & \GG{87.93}{+0.60} & \GG{89.87}{+0.34} & \GG{88.83}{+0.32} & \GG{87.76}{+0.59} & \GG{86.69}{+0.42} \\ 
        \midrule[0.4pt]
        \multirow{2}{*}{Ovis1.6-Gemma2} &
       \textit{Original} & 87.96 & 86.88 & 88.96 & 87.87 & 86.22 & 85.32 \\
        & \textit{+AVP}  & \GG{88.44}{+0.48} & \GG{87.40}{+0.52} & \GG{89.59}{+0.65} & \GG{88.54}{+0.67} & \GG{86.85}{+0.63} & \GG{86.03}{+0.71} \\
        \bottomrule[1.2pt]
        \end{tabular}
    }
    \end{center}
\end{table*}

\subsection{Experimental Results}
\textbf{Results on text-axis hallucination evaluation} 
\Cref{st:SOTA_Comparison} presents the comparative results under the POPE (Polling-based Object Probing Evaluation) evaluation setting~\footnote{Due to space limitations, complete precision and recall results are provided in \Cref{sec:appendix_results_recall_precision}.}.
Our experimental methodology encompasses three distinct sampling strategies: Random Sampling, Popular Sampling, and Adversarial Sampling for negative object sampling, with each strategy generating 3,000 evaluation triplets. 
Across all sampling settings, the integration of VAP through visual noise injection consistently improved the performance of eight state-of-the-art LVMs, with the most substantial gains observed in Intern-VL2, achieving improvements of +2.81\% in accuracy and +2.09\% in F1 score. 
Notably, the most significant improvements were observed under adversarial sampling conditions (\Cref{fig:teaser}-right), indicating that VAP effectively mitigates parametric knowledge bias in LVMs. This is particularly relevant as adversarial sampling tends to generate high-frequency hallucination objects, thereby highlighting the inherent data distribution bias in LVM training sets and the predominant role of LLMs.

\begin{table*}[t]
    \begin{center}
    \caption{Vision-/text-Axis evaluation comparison under the BEAF Benchmark.
    Compared to the text-axis hallucination evaluation, BEAF includes the change-aware hallucination metrics: TU, IG, SB$_p$, SB$_n$, ID, and F1$_{\text{TUID}}$. Although some metrics show slight degradation, the overall performance demonstrates consistent improvement. The values in green indicate the percentage improvements achieved by our proposed method, while the values in red reflect the performance degradation.}
    \vspace{-0em}
    \label{st:SOTA_BEAF}
    \renewcommand{\arraystretch}{1}  % Increase row spacing
    \resizebox{\textwidth}{!}
    {
        \begin{tabular}{lcllllllll}
        \toprule[1.2pt] 
        \multicolumn{1}{l}{\multirow{2}{*}{\textbf{LVM}}} & \multicolumn{1}{c}{\multirow{2}{*}{\textbf{Vision Input}}} & \multicolumn{8}{c}{\textbf{BEAF Benchmark}} \\
        \cmidrule[0.6pt](lr){3-10}
         & &\textbf{Acc.}$\uparrow$ &\textbf{F1}$\uparrow$ & \textbf{TU}$\uparrow$ & \textbf{IG}$\downarrow$ & \textbf{SB}$_\text{p}$$\downarrow$ & \textbf{SB}$_\textbf{n}$$\downarrow$ & \textbf{ID} $\downarrow$ & \textbf{F1}$_\text{TUID}$ $\uparrow$  \\ 
        \midrule[0.8pt]
        \multicolumn{1}{l}{\multirow{2}{*}{LLaVA-v1.5}}  
        & \textit{Original}  &79.99&74.06&34.25&0.33&60.74&4.66&5.42&50.31 \\ 
        & \textit{+VAP}   &\GG{80.36}{+0.37}&\GG{74.35}{+0.29}&\GG{34.83}{+0.58}&\GG{0.27}{-0.06}&\GG{60.72}{-0.02}&\GG{4.18}{-0.46}&\GG{5.05}{-0.37}&\GG{50.97}{+0.66} \\ \midrule[0.4pt]
        \multicolumn{1}{l}{\multirow{2}{*}{Instruct-BLIP}}    
        & \textit{Original}  &81.91&73.55&33.35&0.78&50.73&15.12&5.45&49.30 \\ 
        & \textit{+VAP}    &\GG{82.07}{+0.16}&\GG{73.96}{+0.41}&\GG{33.83}{+0.48}&\GG{0.48}{-0.30}&\GG{50.59}{-0.14}&\GG{15.10}{-0.02}&\GG{5.30}{-0.15}&\GG{49.85}{+0.55} \\ \midrule[0.4pt]
        \multicolumn{1}{l}{\multirow{2}{*}{Intern-VL2}}    
        & \textit{Original}  &88.38&79.10&64.12&1.33&12.63&21.89&6.20&76.17 \\ 
        & \textit{+VAP}   &\GG{88.69}{+0.31}&\GG{79.72}{+0.62}&\GG{66.15}{+2.03}&\GG{0.97}{-0.36}&\GG{11.58}{-1.05}&\GG{21.28}{-0.61}&\GG{6.05}{-0.15}&\GG{77.63}{+1.46} \\ \midrule[0.4pt]
        \multicolumn{1}{l}{\multirow{2}{*}{Intern-VL2-MPO}}    
        & \textit{Original}  &89.21&82.56&63.24&0.76&23.67&12.31&5.23&75.86 \\ 
        & \textit{+VAP}   &\GG{89.63}{+0.42}&\GG{82.72}{+0.18}&\GG{65.06}{+1.78}&\GG{0.45}{-0.31}&\GG{21.91}{-1.76}&12.55\R{+0.24}&\GG{4.49}{-0.74}&\GG{77.40}{+1.66} \\ \midrule[0.4pt]
        \multicolumn{1}{l}{\multirow{2}{*}{DeepSeek-VL2}}    
        & \textit{Original}  &89.39&82.51&67.04&0.50&17.88&14.56&3.02&79.27 \\ 
        & \textit{+VAP}   &\GG{89.72}{+0.33}&\GG{83.12}{+0.61}&\GG{68.11}{+1.07}&\GG{0.44}{-0.06}&\GG{17.37}{-0.51}&\GG{14.06}{-0.50}&\GG{2.98}{-0.04}&\GG{80.03}{+0.76} \\ \midrule[0.4pt]
        \multicolumn{1}{l}{\multirow{2}{*}{Qwen-VL2}}    
        & \textit{Original}  &87.96&81.13&54.78&0.28&33.68&11.24&4.89&69.78 \\ 
        & \textit{+VAP}   &\GG{88.39}{+0.43}&\GG{81.57}{+0.44}&\GG{56.18}{+1.40}&\GG{0.27}{-0.01}&\GG{32.49}{-1.19}&\GG{11.03}{-0.21}&\GG{4.38}{-0.51}&\GG{70.79}{+1.01} \\ \midrule[0.4pt]
        \multicolumn{1}{l}{\multirow{2}{*}{LLaVA-OV}}    
        & \textit{Original}  &90.76&84.53&65.80&0.12&21.32&12.77&2.55&78.56 \\ 
        & \textit{+VAP}    &\GG{91.07}{+0.33}&\GG{85.01}{+0.48}&\GG{67.16}{+1.36}&0.30\R{+0.18}&\GG{20.81}{-0.51}&\GG{11.73}{-1.04}&\GG{2.46}{-0.09}&\GG{79.54}{+0.98} \\ \midrule[0.4pt]
        \multicolumn{1}{l}{\multirow{2}{*}{Ovis1.6-Gemma2}}    
        & \textit{Original}  &90.12&83.04&66.25&0.28&19.94&13.52&2.76&78.80 \\ 
        & \textit{+VAP}    &\GG{90.91}{+0.79}&\GG{84.53}{+1.49}&\GG{68.56}{+2.31}&\GG{0.25}{-0.03}&\GG{19.69}{-0.25}&\GG{11.48}{-2.04}&\GG{2.41}{-0.25}&\GG{80.54}{+1.74} \\ 
        \bottomrule[1.2pt]
        \end{tabular}
        }
    \end{center}
\end{table*}
\textbf{Results on vision-/text-axis hallucination evaluation} 
\Cref{st:SOTA_BEAF} presents the comparative results under the BEAF (BEfore-AFter) evaluation framework. Compared to POPE, BEAF offers superior manipulation along the vision axis and introduces a change-aware metric, providing more insight than standard accuracy in single-scene evaluations. Following the application VAP, all LVMs demonstrated consistent performance improvements across most metrics, with only minor degradations in special cases, which do not detract from the overall efficacy of our method.
Specifically, the most substantial performance improvement for TU was 2.31\%, SB$_p$ improved by 1.76\%, SB$_n$ increased by 1.04\%, and  F1$_{\text{TUID}}$ demonstrated an improvement of 1.74\%. 

The performance improvements in TU, IG, SB$_p$, SB$_n$, ID, and  F1$_{\text{TUID}}$ suggest that our method effectively mitigates hallucinations under varying scene conditions, demonstrating a genuine understanding of object presence beyond reliance on spurious correlations from LVMs' parametric biases or language priors. Importantly, the introduction of VAP to the original images markedly enhanced the TU metric, indicating that the noise added to visual inputs is beneficial. This perturbation aids LVMs in making clearer decisions, thereby reducing confusion and advancing them towards true intelligence~\cite{ye2024beaf}. This further corroborates the effectiveness of our adversarial strategies in suppressing parametric knowledge bias, highlighting their validity.

\begin{table}[t]
    \begin{center}
    \caption{Comparison of object hallucination evaluation under the CHAIR setting. $\boldsymbol{I_1}$ denotes ``\textit{Generate a short caption of the image}'', and $\boldsymbol{I_2}$ denotes ``\textit{Provide a brief description of the given image}''. The values in green indicate the percentage improvements achieved by our proposed method.}
    \vspace{-0em}
    \label{st:SOTA_chair}
    \resizebox{\linewidth}{!}
    {
        \begin{tabular}{lcllll}
        \toprule[1.2pt] 
        \multicolumn{1}{l}{\multirow{2}{*}{\textbf{LVM}}} & \multicolumn{1}{c}{\multirow{2}{*}{\textbf{Vision Input}}} & \multicolumn{2}{c}{$\boldsymbol{I_1}$} & \multicolumn{2}{c}{$\boldsymbol{I_2}$}  \\
        \cmidrule(lr){3-4} \cmidrule(lr){5-6}
         & &\textbf{ CHAIR$_I$} $\downarrow$ &\textbf{ CHAIR$_S$} $\downarrow$ & \textbf{CHAIR$_I$} $\downarrow$ & \textbf{CHAIR$_S$} $\downarrow$  \\ 
        \midrule[0.8pt]
        \multicolumn{1}{l}{\multirow{2}{*}{LLaVA-v1.5}}
        & \textit{Original} &3.97&6.60&4.01&6.90 \\
        & \textit{+VAP}   &\GG{3.82}{-0.15}&\GG{6.50}{-0.10}&\GG{3.86}{-0.15}&\GG{6.50}{-0.40} \\ \midrule[0.4pt]
        \multicolumn{1}{l}{\multirow{2}{*}{Instruct-BLIP}} 
        & \textit{Original} &1.83&2.90&2.14&3.40 \\
        & \textit{+VAP}   &\GG{1.71}{-0.12}&\GG{2.70}{-0.20}&\GG{1.96}{-0.18}&\GG{3.10}{-0.30} \\ \midrule[0.4pt]
        \multicolumn{1}{l}{\multirow{2}{*}{Intern-VL2}} 
        & \textit{Original} &4.90&7.50&5.14&9.50 \\
        & \textit{+VAP}   &\GG{4.22}{-0.68}&\GG{6.60}{-0.90}&\GG{4.65}{-0.49}&\GG{8.90}{-0.60} \\ \midrule[0.4pt]
        \multicolumn{1}{l}{\multirow{2}{*}{Intern-VL2-MPO}} 
        & \textit{Original} &5.53&8.90&6.35&13.40 \\
        & \textit{+VAP}   &\GG{5.39}{-0.14}&\GG{8.60}{-0.30}&\GG{6.17}{-0.18}&\GG{12.60}{-0.80} \\ \midrule[0.4pt]
        \multicolumn{1}{l}{\multirow{2}{*}{DeepSeek-VL2}} 
        & \textit{Original} &2.00&2.60&1.84&4.50 \\
        & \textit{+VAP}   &\GG{1.94}{-0.06}&\GG{2.20}{-0.40}&\GG{1.66}{-0.18}&\GG{4.30}{-0.20} \\ \midrule[0.4pt]
        \multicolumn{1}{l}{\multirow{2}{*}{Qwen-VL2}} 
        & \textit{Original} &3.27&5.20&3.45&6.20 \\
        & \textit{+VAP}   &\GG{2.98}{-0.29}&\GG{4.80}{-0.40}&\GG{3.23}{-0.22}&\GG{5.70}{-0.50} \\ \midrule[0.4pt]
        \multicolumn{1}{l}{\multirow{2}{*}{LLaVA-OV}} 
        & \textit{Original} &1.96&3.30&2.71&4.50 \\
        & \textit{+VAP}   &\GG{1.85}{-0.11}&\GG{3.10}{-0.20}&\GG{2.41}{-0.30}&\GG{4.20}{-0.30} \\ \midrule[0.4pt]
        \multicolumn{1}{l}{\multirow{2}{*}{Ovis1.6-Gemma2}} 
        & \textit{Original} &4.07&6.30&5.80&14.50 \\
        & \textit{+VAP}   &\GG{3.90}{-0.17}&\GG{6.20}{-0.10}&\GG{5.56}{-0.24}&\GG{14.30}{-0.20} \\
        \bottomrule[1.2pt]
        \end{tabular}
    }
    \end{center}
\end{table}
\textbf{Results on open-end caption generation hallucination evaluation}
\Cref{st:SOTA_chair} presents the results of our model under the CHAIR~(Caption Hallucination Assessment with Image Relevance) setting~\footnote{CHAIR restricts object recognition to 80 segmentation categories, which may introduce classification biases in LVMs~\cite{pope}. We limit the response to 30 characters to evaluate hallucination on the most prominent objects.}. 
Upon applying optimized VAP to the original images, we observed significant performance improvements across diverse query prompts, consistently mitigating object hallucination. For instance, under the query prompt ``Generate a short caption of the image'', Intern-VL2 demonstrated reductions of 0.68 and 0.90 in CHAIR$_I$ and CHAIR$_S$ under beneficial visual nosise respectively. 

These empirical results demonstrate the versatility of VAP in open-ended visual language tasks beyond traditional yes/no binary evaluation. By effectively mitigating object hallucination, our approach enhances the reliability and accuracy of complex caption generation, which is essential for applications requiring precise and contextually appropriate descriptions. VAP optimizes semantic alignment between responses and visual content, ensuring generated captions accurately portray salient image features. Additionally, it reduces inherent parametric knowledge bias in LVMs, resulting in generated captions that are both contextually relevant and semantically correct.

% \begin{figure*}[t]
%     \centering
%     \begin{tabular}{@{}c@{\hspace{-0.25mm}}c@{}}
%         \includegraphics[width=0.495\linewidth]{Figure/Accuracy_Comparison_ICML.pdf}&
%         \includegraphics[width=0.495\linewidth]{Figure/F1_Score_Comparison_ICML.pdf}\\
%         % \parbox{0.495\linewidth}{\centering \footnotesize{\ \  (a)}} & 
%         % \parbox{0.495\linewidth}{\centering \footnotesize{\ \  (b)}}
%     \end{tabular}
%     \vspace{-0em}
%     \caption{Comparison of the original images with our proposed VAP and Gaussian noise of equal strength~($\epsilon=2$). We highlight the performance degradation when adding Gaussian noise compared to VAP. The experiments were conducted under the POPE adversarial evaluation setting, with evaluations on Accuracy and F1 Score.}
%     \label{fig:gaussian_vap}
%     \vspace{-0.0em}
% \end{figure*}

\subsection{Analysis and Discussion}

\begin{figure}[t]
        \begin{center}
        \includegraphics[width=\linewidth]{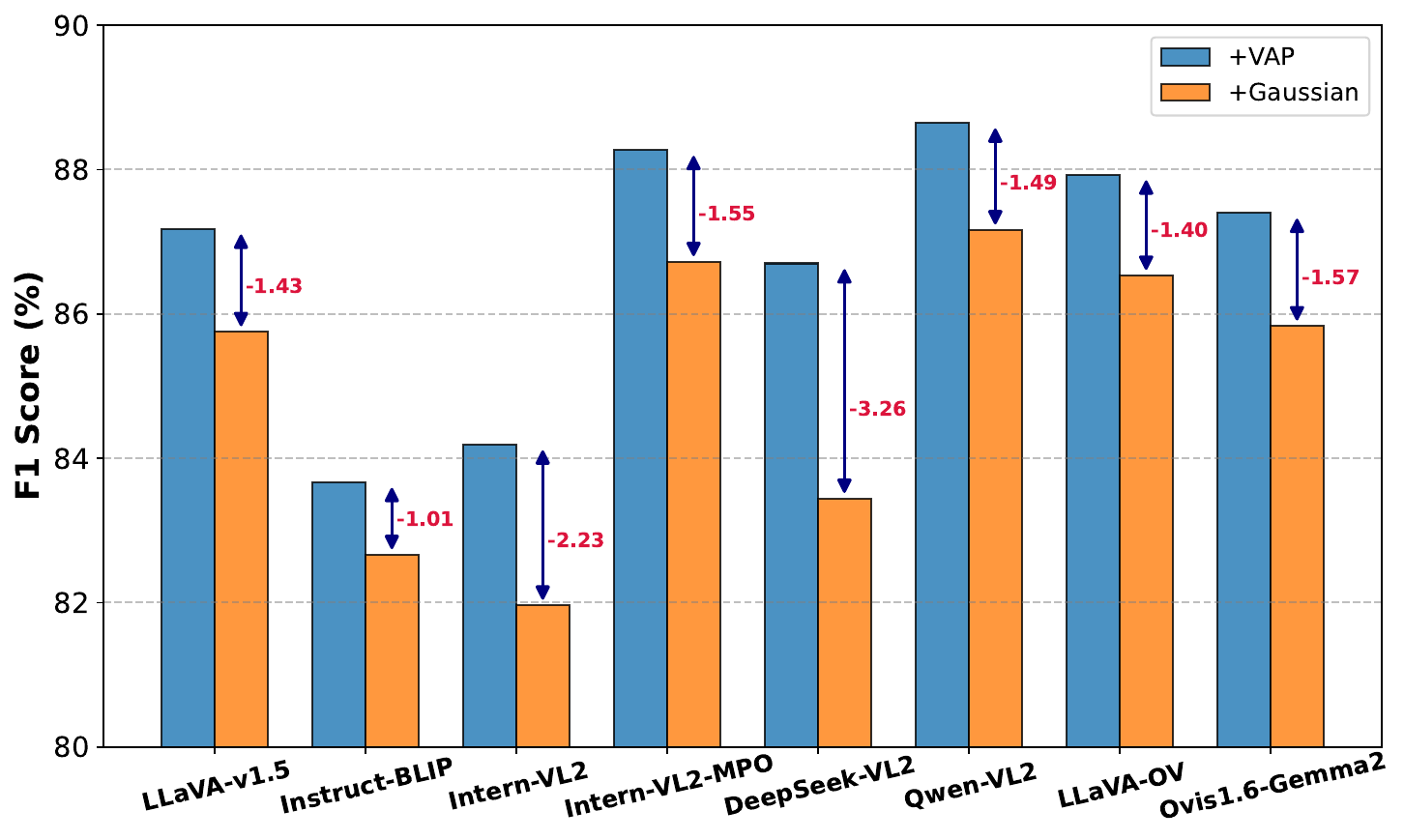}
        \end{center}
        \caption{Comparison of the original images with our proposed VAP and Gaussian noise of equal strength~($\epsilon=2$).  We highlight the performance degradation when adding Gaussian noise compared to VAP. The experiments were conducted using eight SOTA LVMs under the POPE popular evaluation setting, with evaluations on F1 Score.}
        \label{fig:gaussian_vap}
\end{figure}

\begin{figure*}[t]
        \begin{center}
        \includegraphics[width=\linewidth]{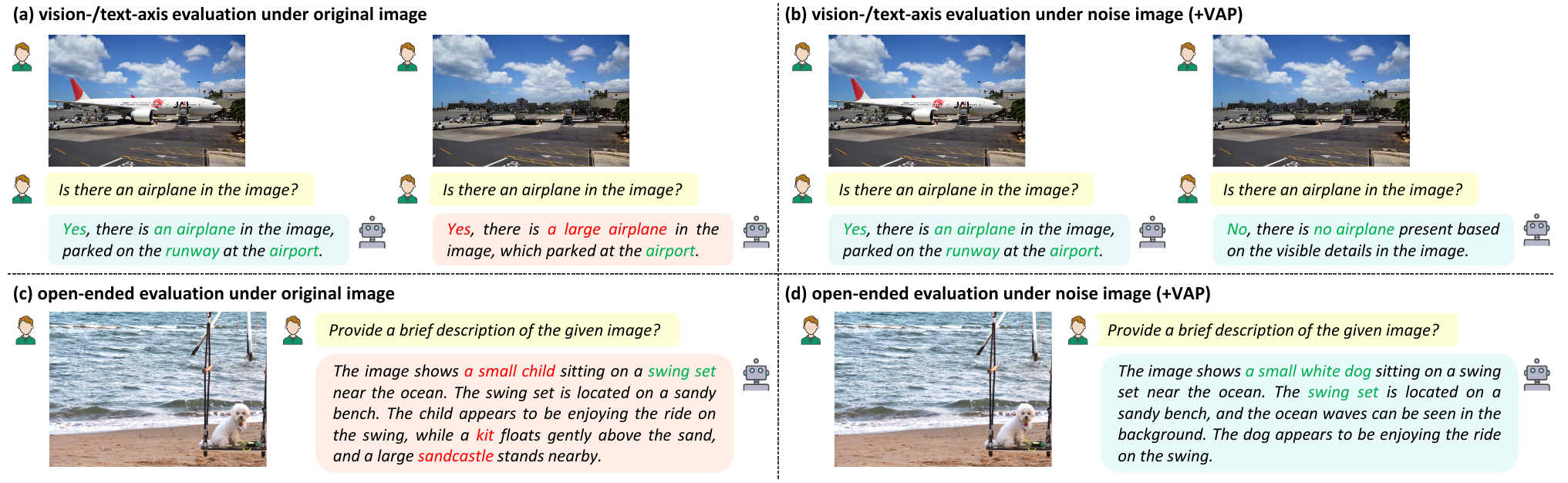}
        \end{center}
        \caption{Examples of the vision-question-answer~(VQA) tasks before and after applying our proposed method to the original images. (a) and (b) demonstrates the suppression of hallucinations in vision-/text-axis evaluations. (c) and (d) shows the reduction of hallucinations in open-ended tasks. Specifically, we use the LLaVA-v1.5~\cite{llava} as an example.}
        \label{fig:example}
\end{figure*}

\begin{figure}[t]
        \begin{center}
        \includegraphics[width=\linewidth]{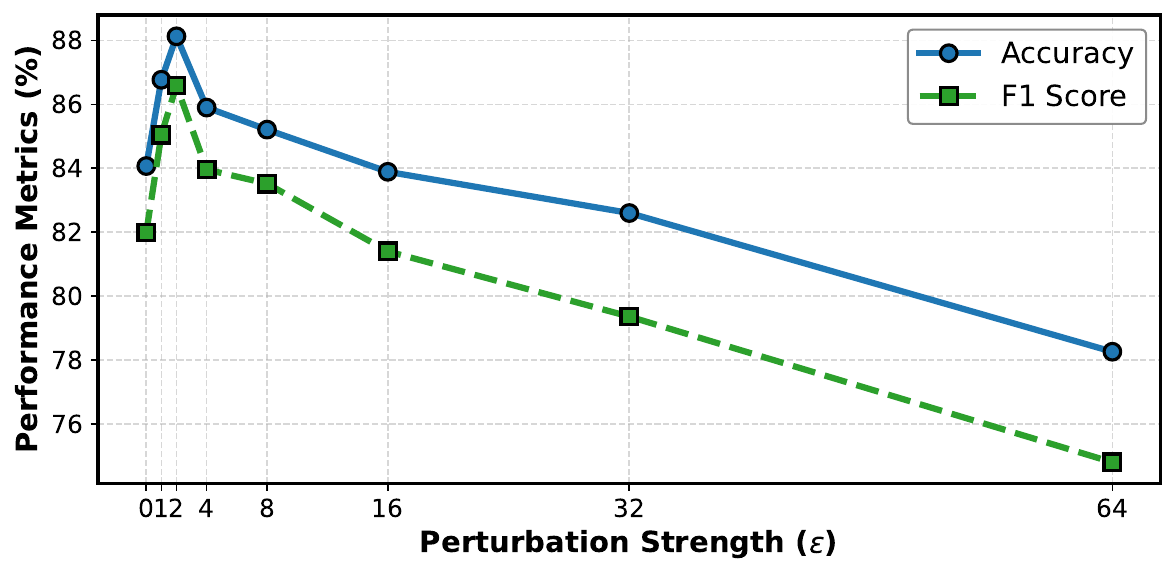}
        \end{center}
        \caption{Performance of the Intern-VL2 model~\cite{InternVL} under varying levels of perturbation strength in the POPE adversarial setting. We test the model's performance with varying perturbations applied to the original images.}
        \label{fig:strength}
\end{figure}

% \begin{figure}[h]
%     \centering
%     \begin{tabular}{@{}c@{}}
%         \includegraphics[width=0.6\linewidth]{Figure/Accuracy_Comparison_ICML.pdf} \\
%         \includegraphics[width=0.6\linewidth]{Figure/F1_Score_Comparison_ICML.pdf}
%     \end{tabular}
%     \vspace{-0em}
%     \caption{Comparison of the original images with our proposed VAP and Gaussian noise of equal strength~($\epsilon=2$). We highlight the performance degradation when adding Gaussian noise compared to VAP. The experiments were conducted under the POPE adversarial evaluation setting, with evaluations on Accuracy and F1 Score.}
%     \label{fig:gaussian_vap}
%     \vspace{-0.0em}
% \end{figure}

\textbf{Effectiveness of VAP and Gaussian noise on hallucinations} 
In \Cref{fig:gaussian_vap}, we compare the performance of adding VAP versus standard Gaussian noise to original images. We observed that, under equally intense perturbations, Gaussian noise consistently and significantly degrades performance across eight models compared to VAP. This substantiates VAP's effectiveness in three ways:
Firstly, VAP introduces beneficial noise, whereas Gaussian noise merely increases visual uncertainty and disrupts visual features.
Secondly, despite equally intense perturbations, VAP optimizes semantic alignment between the model's outputs and visual content through its adversarial strategy, mitigating object hallucination.
Thirdly, unlike Gaussian noise, which only obscures image clarity without aiding model inference, VAP alleviates object hallucination by introducing noise that effectively challenges the model's decision-making process semantically to reduce parametric knowledge bias.

\textbf{Impact of visual adversarial perturbation strength}
\Cref{fig:strength} illustrates the performance variations of our method under different perturbation strengths ($\epsilon$). It can be observed that the model's performance initially improves compared to the scenario without VAP, reaching a peak before declining. When $\epsilon \geq 16$, the performance drops below the baseline without VAP. This indicates that, firstly, our method is effective in mitigating model hallucinations. Secondly, the perturbation strength must not be excessive, as overly strong perturbations can disrupt visual features, leading to a decline in model performance. 

\textbf{Illustration of the effectiveness on closed VQA and open-ended tasks}
\Cref{fig:strength} presents results from specific examples in closed vision-question-answer (VQA) and open-ended image captioning tasks. Panels (a) and (b) demonstrate that the visual noise introduced by our method effectively suppresses object hallucinations in LVMs under scene change situations, without disrupting their normal perceptual capabilities (i.e., the noise does not lead to incorrect decisions). Additionally, Panels (c) and (d) further show that our method mitigates object hallucinations in open-ended tasks without reducing the amount of information in the LVMs' responses. These consistent findings highlight the effectiveness of the VAP method.  More comprehensive examples can be found in \Cref{sec:appendix_demo}.
\section{Conclusion}
This paper introduces visual adversarial perturbation (VAP), a novel data-centric method designed to mitigate object hallucinations in large vision-language models (LVMs) by introducing imperceptible noise to visual inputs. Unlike model-centric strategies that require modifying the complex LVMs, VAP strategically introduces beneficial noise to visual data to ground its responses with actual visual content and mitigate reliance on biased parametric knowledge in LVMs. Extensive experiment evaluations across POPE, BEAF, and CHAIR benchmarks demonstrate that VAP effectively reduces object hallucinations in various settings, enhancing the reliability of LVMs. 

These findings underscore the effectiveness of leveraging visual adversarial perturbations as a novel ``poison as cure'' strategy for mitigating object hallucinations, demonstrated for the first time in this work. While this approach proves effective, the optimization of visual noise is computationally intensive. A straightforward solution is to utilize smaller models as proxies for optimization, which can reduce computational costs to one-eighth, as detailed in \Cref{sec:appendix_genelization}. Exploring the generalization of VAP and reducing computational costs are considered key directions for future work.

\newpage

\section*{Impact Statement}
This work advances the responsible development of artificial intelligence systems by enhancing the reliability and trustworthiness of large vision-language models. By addressing the critical issue of hallucinations, our approach has the potential to improve real-world applications across diverse domains. 
The reduced misinformation risk promotes greater public trust in AI technologies while reducing the risks of misinformation and biased content.

\bibliography{example_paper}
\bibliographystyle{icml2025}

%%%%%%%%%%%%%%%%%%%%%%%%%%%%%%%%%%%%%%%%%%%%%%%%%%%%%%%%%%%%%%%%%%%%%%%%%%%%%%%
%%%%%%%%%%%%%%%%%%%%%%%%%%%%%%%%%%%%%%%%%%%%%%%%%%%%%%%%%%%%%%%%%%%%%%%%%%%%%%%
% APPENDIX
%%%%%%%%%%%%%%%%%%%%%%%%%%%%%%%%%%%%%%%%%%%%%%%%%%%%%%%%%%%%%%%%%%%%%%%%%%%%%%%
%%%%%%%%%%%%%%%%%%%%%%%%%%%%%%%%%%%%%%%%%%%%%%%%%%%%%%%%%%%%%%%%%%%%%%%%%%%%%%%
\newpage
\appendix
\onecolumn

\section{More Details of Experiment Setup}
\label{appendix}

\subsection{More Details about Baseline LVMs}
In this study, we comprehensively selecte eight state-of-the-art large vision-language models (LVMs) carefully selected to validate the effectiveness of our proposed method. As illustrated in \Cref{tab:LVM_Info}, our chosen models span critical developments from September 2023 to December 2024, encompassing parameter ranges from 7.1B to 16.1B and integrating advanced language models like Vicuna, Qwen2, and Gemma2 with sophisticated vision encoders such as CLIP, SigLIP, and custom vision transformers.
Our model selection strategy focuses on capturing the latest architectural innovations in addressing hallucination challenges in vision-language understanding. By examining models from leading research initiatives including LLaVA, Instruct-BLIP, Intern-VL, DeepSeek, Ovis, LLaVA-OV and Qwen, we aim to provide a comprehensive hallucination evaluations of current multimodal AI.
\begin{table}[h]
    \begin{center}
    \caption{Detailed information of large vision-language models used in this paper.}
    \vspace{-0.5em}
    \label{tab:LVM_Info}
    \resizebox{0.9\linewidth}{!}
    {
        \begin{tabular}{lcccc}
        \toprule[1.2pt]
        LVM & \# Parameters & Language Model & Vision Model & Released Date\\
        \midrule[0.8pt]
        LLaVA-v1.5~\cite{llava} & 7.1B & Vicuna-7B & CLIP ViT-L/14 & 2023-09 \\
        Instruct-BLIP~\cite{instruct_blip} & 7.9B & Vicuna-7B & ViT-G & 2023-09 \\
        Intern-VL2~\cite{InternVL} & 8.1B & InternLM2.5-7B & InternViT-300M & 2024-07 \\
        Intern-VL2-MPO~\cite{InternVL} & 8.1B & InternLM2.5-7B & InternViT-300M & 2024-11 \\
        DeepSeek-VL2~\cite{deepseek_vl2} & 16.1B & Gemma2-9B & SigLIP-400M & 2024-12 \\
        Qwen-VL2~\cite{Qwen} & 8.3B & Qwen2-7B & ViT-Qwen & 2024-08 \\
        LLaVA-OV~\cite{llava_ov} & 8.0B & Qwen2-7B & SigLIP-400M & 2024-08 \\
        Ovis1.6-Gemma2~\cite{ovis} & 9.4B & Gemma2-9B & SigLIP-400M & 2024-11 \\
        \bottomrule[1.2pt]
        \end{tabular}
    }
    \end{center}
\end{table}

\subsection{More Details about Implementation Details}
We conducted our experiments across eight state-of-the-art vision-language models: LLaVA-v1.5, Instruct-BLIP, Intern-VL2, Intern-VL2-MPO, DeepSeek-VL2, Qwen-VL2, LLaVA-OV, and Ovis1.6-Gemma2. The experiments were performed using NVIDIA RTX 4090 (24GB), A6000 (48GB), and A100 (80GB) GPUs. For the adversarial parameters, we set $\alpha=1/255$, $\beta=8/255$, $N=10$, and $\epsilon=2$ unless otherwise noted,. Model-specific balance parameters are detailed in \Cref{tab:LVM_parameter}. We employ ViT-L/14 as our default CLIP text encoder ($g_\psi$) unless otherwise specified.

\begin{table}[h]
    \begin{center}
    \caption{Detailed specifications of large vision-language models used in this paper.}
    \vspace{-0.5em}
    \label{tab:LVM_parameter}
    \resizebox{0.7\linewidth}{!}
    {
        \begin{tabular}{lcccc}
        \toprule[1.2pt]
        LVM & $\sqrt{1/\sigma{_1}^2}$ & $\sqrt{1/\sigma{_2}^2}$ & $\sqrt{1/\sigma{_3}^2}$ & $T$ \\
        \midrule[0.8pt]
        LLaVA-v1.5~\cite{llava} & 1.0 & 1.0 & 1.0 & 500 \\
        Instruct-BLIP~\cite{instruct_blip} & 1.0 & 1.0 & 1.0 & 500 \\
        Intern-VL2~\cite{InternVL} & 1.0 & 0.5 & 0.5 & 200 \\
        Intern-VL2-MPO~\cite{InternVL} & 1.0 & 0.5 & 0.5 & 800 \\
        DeepSeek-VL2~\cite{deepseek_vl2} & 1.0 & 1.0 & 1.0 & 100 \\
        Qwen-VL2~\cite{Qwen} & 1.0 & 0.5 & 0.5 & 500 \\
        LLaVA-OV~\cite{llava_ov} & 0.1 & 1.0 & 0.1 & 200 \\
        Ovis1.6-Gemma2~\cite{ovis} & 1.0 & 1.0 & 1.0 & 500 \\
        \bottomrule[1.2pt]
        \end{tabular}
    }
    \end{center}
\end{table}

\section{More Details of Evaluation Benchmark}
\label{sec:appendix_evaluation}
\subsection{POPE Evaluation}
POPE (Polling-based Object Probing Evaluation)~\cite{pope} is a simple yet effective framework for assessing object hallucinations in LVMs. POPE formulates the evaluation of object hallucinations as a series of binary (yes/no) classification tasks. By sampling hallucinated objects, POPE constructs triplets of the form:
\begin{equation}
    \langle x,c,w_{(gt)} \rangle,
\end{equation}
where \( x \) represents the queried image, \( c \) is the query prompt template, and \( w_{(gt)} \) is the ground-truth answer to the query. The triplets generated by POPE include those with a ``yes'' response based on ground-truth objects and ``no'' responses obtained by sampling from negative objects. There are three strategies for negative sampling:
\begin{itemize}
    \item \textbf{Random Sampling}: Randomly samples objects that do not exist in the image.
    \item \textbf{Popular Sampling}: Selects the top-\( k \) most frequent objects in the image dataset that are absent from the current image.
    \item \textbf{Adversarial Sampling}: Ranks all objects based on their co-occurrence frequencies with the ground-truth objects and selects the top-\( k \) frequent ones that do not exist in the image.
\end{itemize}

POPE employs the following evaluation metrics to measure performance:

\begin{gather}
    \text{Accuracy} = \frac{\text{TP}+\text{TN}}{\text{TP}+\text{TN}+\text{FP}+\text{FN}}, \\
    \text{Precision} = \frac{\text{TP}}{\text{TP} + \text{FP}}, \\
    \text{Recall} = \frac{\text{TP}}{\text{TP} + \text{FN}}, \\
    \text{F1 Score} = 2 \times \frac{\text{Precision} \times \text{Recall}}{\text{Precision} + \text{Recall}}.
\end{gather}

In the above equations:
\begin{itemize}
\item \textbf{TP (True Positives)}: The number of correctly identified objects that are present in the image.
\item \textbf{TN (True Negatives)}: The number of correctly identified objects that are absent from the image.
\item \textbf{FP (False Positives)}: The number of objects incorrectly identified as present in the image.
\item \textbf{FN (False Negatives)}: The number of objects that are present in the image but were not identified by the model.
\end{itemize}

These metrics provide a comprehensive evaluation of the model's ability to accurately identify the presence or absence of objects, thereby quantifying the extent of hallucinations in LVMs.

\subsection{BEAF Evaluation}

BEAF (BEfore and AFter)~\cite{ye2024beaf} extends the evaluation framework beyond the text-axis hallucination assessment of POPE by simultaneously considering both text- and vision-axes. Additionally, BEAF introduces change-aware metrics, enabling a more granular evaluation of object hallucinations. Similar to POPE, BEAF employs binary classification tasks using triplets; however, it accounts for more complex perceptual changes within the dataset.

\textbf{Dataset Definition}
BEAF utilizes a dataset \( G \) composed of tuples:
\begin{equation}
    G = \{(X_o, X_m, C, W_o, W_m, E)\}_{i=1}^{|G|},
\end{equation}
where \( X_o \) denotes the original image. \( X_m \) represents the change-aware manipulate image. \( C \) is the question. \( W_o \) and \( W_m \) are the corresponding answers for the original and manipulated images, respectively. \( E \in \{\text{True}, \text{False}\} \) indicates whether the question pertains to an object that has been removed in the manipulated image.

\textbf{Filter Function}
To facilitate the extraction of specific subsets from \( G \) based on input conditions, BEAF defines a filter function:
\begin{equation}
    \text{Filter}(b_o, b_m, b_r) = \{ h \mid \text{IsCorrect}(W_o) = bo, \, \text{IsCorrect}(W_m) = b_m, \, E = b_r, \, h \in G \},
\end{equation}
where \( h = (X_o, X_m, C, W_o, W_m, E) \). Here, \( b_o \), \( b_m \), and \( b_r \) are boolean values $\{\text{True}, \text{False}\}$ that specify the desired correctness and relation flags for filtering.

\textbf{Evaluation Metrics}
Based on the \texttt{Filter} function, BEAF defines the following fine-grained perceptual change metrics:

\begin{gather}
    \text{TU} = \frac{|\text{Filter(True, True, True)}|}{|\text{Filter(True $\lor$ False, True $\lor$ False, True)}|} \times 100, \\
    \text{IG} = \frac{|\text{Filter(False, False, True)}|}{|\text{Filter(True $\lor$ False, True $\lor$ False, True)}|} \times 100, \\
    \text{SB}_p = \frac{|\text{Filter(True, False, True)}|}{|\text{Filter(True $\lor$ False, True $\lor$ False, True)}|} \times 100, \\
    \text{SB}_n = \frac{|\text{Filter(False, True, True)}|}{|\text{Filter(True $\lor$ False, True $\lor$ False, True)}|} \times 100, \\
    \text{ID} = \frac{|\text{Filter(True, False, False)}| + |\text{Filter(False, True, False)}|}{|\text{Filter(True $\lor$ False, True $\lor$ False, False)}|} \times 100, \\
    \text{F1}_{\textbf{TUID}} = \frac{2 \times \text{TU}}{1 + (100 - \text{ID})},
\end{gather}
where TU represents True Understanding, IG denotes Ignorance, SB refers to Stubbornness, and ID signifies Indecision. These metrics provide a more nuanced evaluation of the model's capacity to recognize and adapt to perceptual changes across textual and visual contexts, offering a comprehensive assessment of hallucinations in LVMs.

\section{More Details of Experiment Results}
\label{sec:appendix_results}
\subsection{Evaluation of Text-Axis and Vision-/Text-Axis Hallucinations}
\label{sec:appendix_results_recall_precision}
\Cref{st:SOTA_Comparison_pre_recall} presents the performance evaluation of Precision (Prec.) and Recall under the POPE and BEAF experimental settings. The results demonstrate that our method achieves effective improvements in both text-axis and vision-/text-axis hallucination evaluations. While a slight decrease in Recall is observed in some cases, the overall performance exhibits significant enhancement. Notably, the decline in Recall is minimal, whereas the improvement in Precision is more pronounced, further validating the effectiveness of our approach.
\begin{table*}[htb]
    \begin{center}
    \caption{Comparison of text-axis evaluation across three POPE evaluation settings: Random Sampling, Popular Sampling, and Adversarial Sampling on the MSCOCO validation set. Additionally, vision- and text-axis evaluations are conducted under the BEAF benchmark. The values highlighted in green represent the percentage improvements achieved by our proposed method, whereas the values in red indicate performance degradation.}
    \vspace{-0em}
    \label{st:SOTA_Comparison_pre_recall}
    \resizebox{\textwidth}{!}
    {
        \begin{tabular}{lcllllllll}
        \toprule[1pt] 
        \multicolumn{1}{l}{\multirow{2}{*}{\textbf{LVM}}} & \multicolumn{1}{c}{\multirow{2}{*}{\textbf{Vision Input}}} & \multicolumn{2}{c}{\textbf{POPE-Popular}} & \multicolumn{2}{c}{\textbf{POPE-Random}} & \multicolumn{2}{c}{\textbf{POPE-Adversarial}} & \multicolumn{2}{c}{\textbf{BEAF}} \\
        \cmidrule(lr){3-4} \cmidrule(lr){5-6} \cmidrule(lr){7-8} \cmidrule(lr){9-10}
         & & \textbf{Prec.$\uparrow$} & \textbf{Recall$\uparrow$} & \textbf{Prec.$\uparrow$} & \textbf{Recall$\uparrow$} & \textbf{Prec.$\uparrow$} & \textbf{Recall$\uparrow$} & \textbf{Prec.$\uparrow$} & \textbf{Recall$\uparrow$} \\ 
        \midrule
        \multicolumn{1}{l}{\multirow{2}{*}{LLaVA-v1.5}}
        & \textit{Original} &82.87&90.09&88.13&90.07&74.45&90.73&61.77&92.43 \\
        & \textit{+VAP}  &\GG{83.95}{+1.08}&\GG{90.67}{+0.58}&\GG{89.47}{+1.34}&\GG{90.67}{+0.60}&\GG{75.27}{+0.82}&\GG{92.04}{+1.31}&\GG{62.32}{+0.55}&92.13\R{-0.30} \\ \midrule
        \multicolumn{1}{l}{\multirow{2}{*}{Instruct-BLIP}} 
        & \textit{Original} &85.15&80.67&94.83&80.67&82.21&81.33&67.00&81.52 \\
        & \textit{+VAP}  &\GG{85.78}{+0.63}&\GG{81.67}{+1.00}&\GG{95.70}{+0.87}&\GG{81.67}{+1.00}&\GG{82.50}{+0.29}&\GG{82.42}{+1.09}&\GG{67.47}{+0.47}&\GG{81.83}{+0.31} \\ \midrule
        \multicolumn{1}{l}{\multirow{2}{*}{Intern-VL2}} 
        & \textit{Original} &95.62&71.90&97.40&71.71&92.50&71.64&87.40&72.24 \\
        & \textit{+VAP}  &\textbf{97.41}\G{+1.59}&\textbf{74.13}\G{+2.23}&\textbf{98.07}\G{+0.67}&\textbf{73.58}\G{+1.87}&\textbf{94.50}\G{+2.00}&\textbf{73.66}\G{+2.02}&\GG{88.76}{+1.36}&\GG{72.35}{+0.09} \\ \midrule
        \multicolumn{1}{l}{\multirow{2}{*}{Intern-VL2-MPO}} 
        & \textit{Original} &93.70&80.39&95.39&80.95&90.55&81.08&82.46&82.67 \\
        & \textit{+VAP}  &\GG{94.11}{+0.41}&\GG{83.12}{+2.73}&\GG{96.48}{+1.09}&\GG{83.12}{+2.17}&\GG{91.62}{+1.07}&\GG{83.83}{+2.75}&\GG{83.52}{+1.06}&\GG{82.73}{+0.06} \\ \midrule
        \multicolumn{1}{l}{\multirow{2}{*}{DeepSeek-VL2}} 
        & \textit{Original} &92.46&80.13&96.70&80.13&91.06&80.67&84.11&80.90 \\
        & \textit{+VAP}  &\GG{93.52}{+1.06}&\GG{80.80}{+0.67}&\GG{97.34}{+0.64}&\GG{80.81}{+0.68}&\GG{92.39}{+1.33}&\GG{80.93}{+0.26}&\GG{85.12}{+1.01}&\GG{81.21}{+0.31} \\ \midrule
        \multicolumn{1}{l}{\multirow{2}{*}{Qwen-VL2}} 
        & \textit{Original} &91.15&84.47&96.28&84.47&87.21&84.87&78.62&83.81 \\
        & \textit{+VAP}  &\GG{92.34}{+1.19}&\GG{85.26}{+0.79}&\GG{97.39}{+1.11}&\GG{84.60}{+0.13}&\GG{88.87}{+1.66}&\GG{85.25}{+0.38}&\GG{80.03}{+1.41}&83.14\R{-0.67} \\ \midrule
        \multicolumn{1}{l}{\multirow{2}{*}{LLaVA-OV}} 
        & \textit{Original} &95.20&80.67&98.06&80.67&92.72&80.67&87.58&81.69 \\
        & \textit{+VAP}  &\GG{96.97}{+1.77}&\GG{80.81}{+0.14}&\GG{99.00}{+0.94}&80.56\R{-0.11}&\GG{93.54}{+0.82}&\GG{81.13}{+0.46}&\GG{88.17}{+0.59}&\GG{82.06}{+0.37} \\ \midrule
        \multicolumn{1}{l}{\multirow{2}{*}{Ovis1.6-Gemma2}} 
        & \textit{Original} &95.45&79.72&97.87&79.65&91.19&80.16&86.17&80.95 \\
        & \textit{+VAP}  &\GG{96.74}{+0.29}&79.70\R{-0.02}&\GG{98.44}{+0.57}&\GG{80.45}{+0.80}&\GG{91.69}{+0.50}&\GG{81.03}{+0.87}&\GG{86.92}{+0.75}&\GG{82.27}{+1.32} \\
        \bottomrule[1pt]
        \end{tabular}
    }
    \end{center}
\end{table*}

\subsection{Parameter Sensitive Analysis}
\Cref{st:paramter_analysis} presents the parameter sensitivity analysis of the adversarial strategies loss function, as the parameters used in our approach vary across different models due to their distinct characteristics. The results indicate that parameter choices significantly impact performance metrics, including Accuracy (Acc.), Precision (Prec.), Recall (Rec.), and F1-score (F1). Notably, the selection of $\sqrt{1/\sigma_1}$, $\sqrt{1/\sigma_2}$, and $\sqrt{1/\sigma_3}$ involves a trade-off process, where optimizing one metric may lead to compromises in others. Interestingly, certain parameters yield competitive performance even when set to zero, suggesting potential redundancy in specific configurations. This trade-off underscores the necessity of carefully balancing parameter choices to achieve optimal overall performance.
\begin{table*}[h]
    \begin{center}
    \caption{Parameter analysis of the Intern-VL2~\cite{InternVL} under varying settings of $\sigma_1$, $\sigma_2$, and $\sigma_3$. The model parameters were fixed as $\sqrt{1/\sigma_1}=1.0$, $\sqrt{1/\sigma_2}=0.5$, and $\sqrt{1/\sigma_3}=0.5$ without changing the values of $\sigma_1$, $\sigma_2$, and $\sigma_3$. Performance comparison under the POPE Random evaluation setting, which involves randomly sampling objects that do not exist in the image. We randomly selected 1000 images from the MS-COCO dataset for this evaluation.}
    \label{st:paramter_analysis}
    \renewcommand{\arraystretch}{1}  % Increase row spacing
    \resizebox{\textwidth}{!}
    {
        \begin{tabular}{c| *{12}{l}}
        \toprule[1.2pt]
        \multirow{2}{*}{\textbf{Value}} & 
        \multicolumn{4}{c}{\textbf{$\sqrt{1/\sigma_1}$}} & 
        \multicolumn{4}{c}{\textbf{$\sqrt{1/\sigma_2}$}} & 
        \multicolumn{4}{c}{\textbf{$\sqrt{1/\sigma_3}$}} \\
        \cmidrule[0.5pt](lr){2-5} \cmidrule[0.5pt](lr){6-9} \cmidrule[0.5pt](lr){10-13}
        & \textbf{Acc.~$\uparrow$} & \textbf{Prec.~$\uparrow$} & \textbf{Rec.}~$\uparrow$ & \textbf{F1~$\uparrow$} & 
        \textbf{Acc.~$\uparrow$} & \textbf{Prec.~$\uparrow$} & \textbf{Rec.}~$\uparrow$ & \textbf{F1~$\uparrow$} & 
        \textbf{Acc.~$\uparrow$} & \textbf{Prec.~$\uparrow$} & \textbf{Rec.}~$\uparrow$ & \textbf{F1~$\uparrow$} \\ 
        \midrule[0.8pt]
        0.0    & 87.20 & 95.72 & 77.24 & 85.49 & 86.82 & 95.65 & 76.38 & 84.94 & 87.54 & 94.95 & 78.47 & 85.93 \\ 
        0.1    & 86.77 & 95.61 & 76.22 & 84.82 & 87.75 & 96.52 & 77.62 & 86.04 & 86.82 & 95.65 & 76.38 & 84.94 \\ 
        0.25   & 86.73 & 94.78 & 76.76 & 84.82 & 87.83 & 95.76 & 78.47 & 86.25 & 87.45 & 94.87 & 78.16 & 85.71 \\ 
        0.5    & 87.45 & 95.68 & 77.62 & 85.71 & 88.09 & 96.55 & 78.32 & 86.48 & 87.79 & 95.72 & 78.32 & 86.15 \\ 
        0.75   & 87.24 & 94.95 & 77.93 & 85.60 & 87.83 & 94.95 & 79.02 & 86.25 & 87.58 & 95.79 & 78.08 & 86.03 \\ 
        1.0    & 87.92 & 95.80 & 78.62 & 86.36 & 87.50 & 95.72 & 77.77 & 85.82 & 87.58 & 95.79 & 78.08 & 86.03 \\ 
        \bottomrule[1.2pt]
        \end{tabular}
    }
    \end{center}
\end{table*}

\subsection{Detailed Comparison of VAP and Gaussian Noise}
\begin{figure*}[h]
    \centering
    \begin{tabular}{@{}c@{\hspace{-0.25mm}}c@{}}
        \includegraphics[width=0.495\linewidth]{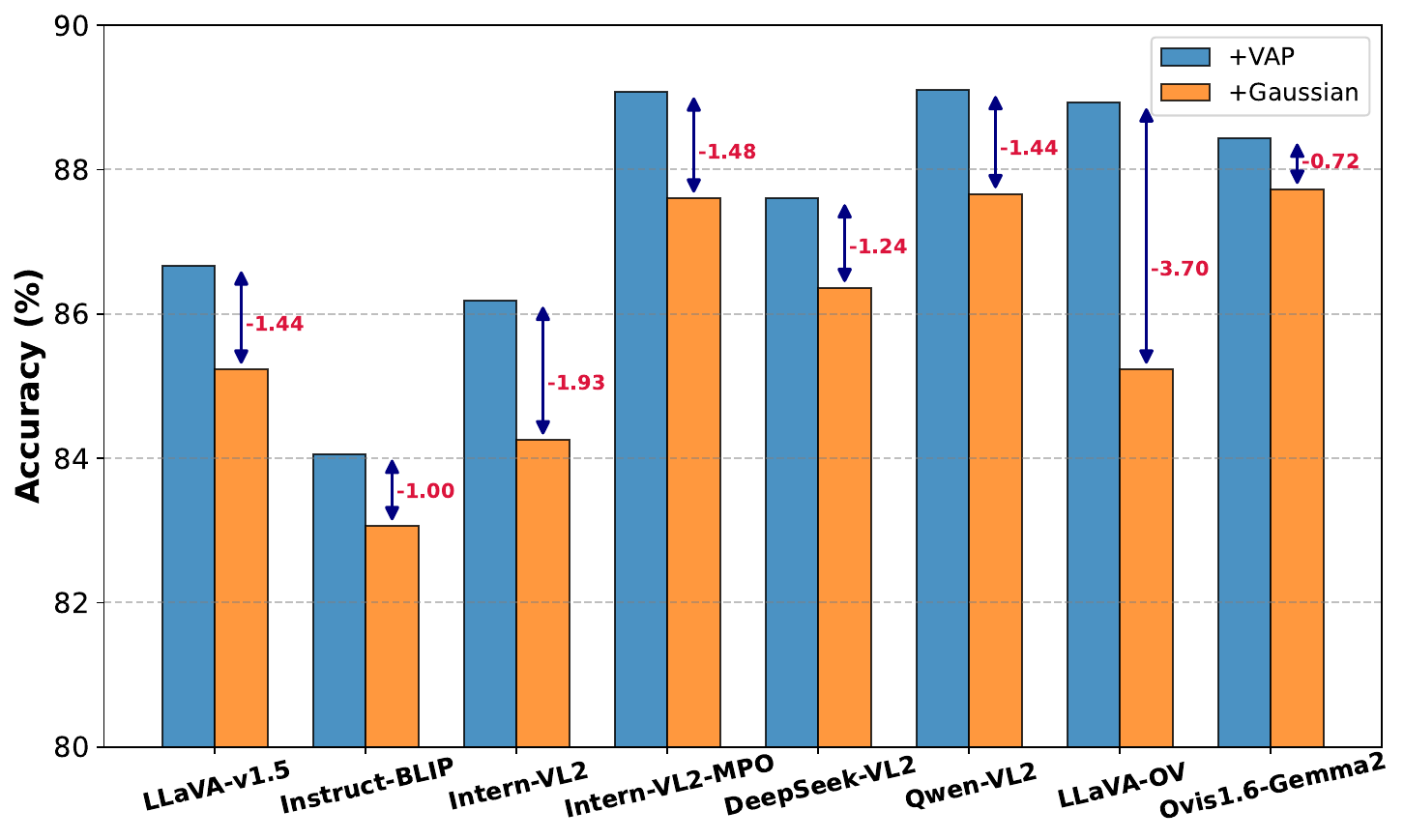}&
        \includegraphics[width=0.495\linewidth]{Figure/F1_Score_Comparison_ICML.pdf}\\
        % \parbox{0.495\linewidth}{\centering \footnotesize{\ \  (a)}} & 
        % \parbox{0.495\linewidth}{\centering \footnotesize{\ \  (b)}}
    \end{tabular}
    \vspace{-0em}
    \caption{Comparison of the original images with our proposed VAP and Gaussian noise of equal strength~($\epsilon=2$). We highlight the performance degradation when adding Gaussian noise compared to VAP. The experiments were conducted under the POPE adversarial evaluation setting, with evaluations on Accuracy and F1 Score.}
    \label{fig:gaussian_vap_all}
    \vspace{-0.0em}
\end{figure*}

\Cref{fig:gaussian_vap_all} comprehensively compares the performance differences observed across eight state-of-the-art LVMs when applying Gaussian noise and VAP with identical intensity to the original images. The results consistently demonstrate that VAP achieves superior performance over Gaussian noise. This consistent improvement highlights the effectiveness of VAP in mitigating hallucinations and reinforcing the model's ability to generate outputs that better align with the visual content.

Unlike Gaussian noise, which introduces random perturbations without a targeted objective, VAP strategically injects adversarial noise designed to align model outputs with visual data. This reduces reliance on spurious correlations and enhances semantic consistency. Furthermore, the observed performance gains across various models underscore VAP's adaptability and robustness, positioning it as a promising data-centric solution for improving LVM reliability.

\subsection{Effect of Different Levels of Visual Uncertainty}
\begin{figure*}[h]
        \begin{center}
        \includegraphics[width=0.65\linewidth]{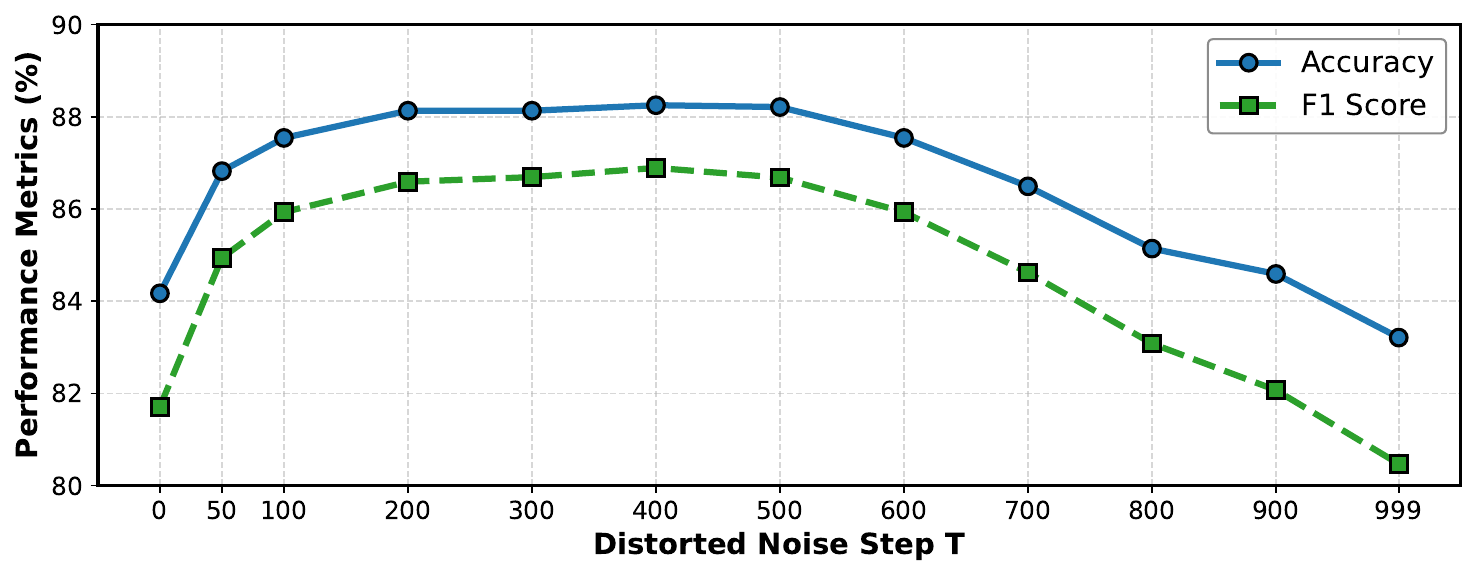}
        \end{center}
        \caption{Performance of the Intern-VL2~\cite{InternVL} under varying levels of distortion strength $T$ in the POPE adversarial setting. The evaluation was conducted on 1000 randomly selected images from the MS-COCO dataset.}
        \label{fig:vraying_T}
\end{figure*}

\Cref{fig:vraying_T} presents the performance variations of LVMs with increasing levels of distortion applied to negative distorted images, as described in \Cref{Distorted_T}. As the distortion level $T$ increases, the model's hallucination initially decreases but subsequently rises, with performance reaching its lowest point when the distorted input consists entirely of Gaussian noise, even lower than when $T=0$. The analysis can be summarized as follows:
(1) Initially, as $T$ increases, hallucinations decrease because the distorted input helps quantify parametric knowledge bias. The VAP method employs a dual-setting approach to reduce the semantic similarity between LVM responses to original and distorted visual inputs under both the conditional setting $c$ (with a query prompt) and the unconditional setting $\emptyset$ (without a query prompt). The optimized visual noise ultimately mitigates parametric knowledge bias in LVMs.
(2) However, when $T$ exceeds a certain threshold, hallucinations increase instead. This occurs because excessive distortion compromises the model's ability to extract meaningful visual information, leading to inaccurate quantification of parametric knowledge bias. 

\section{Generalization of VAP}
\label{sec:appendix_genelization}
The high computational cost of optimizing adversarial strategies poses a significant challenge. A practical approach to mitigate this challenge is to leverage smaller-scale models as proxies to generate visual perturbations. \Cref{st:Generalization} demonstrates the strong generalization capability of VAP, where perturbations generated by smaller models effectively enhance the performance of larger counterparts. Specifically, applying perturbations from the Intern-VL2-1B model to Intern-VL2-8B results in a 1.78\% improvement in F1 score, while substantially reducing inference costs—requiring only $\frac{1}{8}$ of the A100 computation time per sample compared to Intern-VL2-8B. A similar pattern is observed in the Qwen-VL2 series, where proxy-generated noise also leads to consistent performance improvements in larger-scale models. Although the performance gains from proxy-based perturbations are slightly lower than those from target model-generated noise, they provide an effective balance between computational efficiency and performance enhancement. These findings underscore the potential of VAP in scaling hallucination suppression across models of different sizes, offering a scalable and resource-efficient solution for real-world applications.

\begin{table}[h]
    \begin{center}
    \caption{Generalization performance of VAP across different models.  The table compares the results obtained from the original images (left value) and the perturbed images generated using source models under the VAP setting (right value). Experiments are conducted on Intern-VL2 and Qwen-VL2 models, with the best results highlighted in \textbf{bold}. The inference cost reduction, shown in the last row, is measured relative to using the original target models.}
    \vspace{-0em}
    \label{st:Generalization}
    \resizebox{0.85\linewidth}{!}
    {
        \begin{tabular}{lccccc}
        \toprule[1.2pt] 
        \multicolumn{1}{l}{\multirow{2}{*}{\textbf{Metric}}} & \multicolumn{3}{c}{\textbf{Source: Intern-VL2-1B}} & \multicolumn{2}{c}{\textbf{Source: Qwen-VL2-2B}}  \\
        \cmidrule(lr){2-4} \cmidrule(lr){5-6}
         &\textbf{$\Rightarrow$ Intern-VL2-1B}&\textbf{$\Rightarrow$ Intern-VL2-4B}&\textbf{$\Rightarrow$ Intern-VL2-8B}&\textbf{$\Rightarrow$ Qwen-VL2-2B}&\textbf{$\Rightarrow$ Qwen-VL2-7B}  \\ \midrule[0.8pt]
         Accuracy  &81.69/\textbf{83.28}&81.55/\textbf{82.56}&82.00/\textbf{84.07}&84.47/\textbf{85.42}&86.27/\textbf{86.87}\\
         Precision &89.72/\textbf{92.13}&85.65/\textbf{87.21}&87.40/\textbf{90.97}&83.98/\textbf{84.85}&87.21/\textbf{88.03}\\
         Recall    &70.94/\textbf{72.34}&75.05/\textbf{75.90}&72.24/\textbf{75.50}&84.04/\textbf{85.26}&84.87/\textbf{85.33} \\
         F1 Score  &79.23/\textbf{81.04}&80.00/\textbf{81.16}&80.70/\textbf{82.52}&84.01/\textbf{85.05}&86.02/\textbf{86.66}\\
         \midrule[0.8pt]
        Inference Cost Reduction & \textbf{1×} & \textbf{1/3×} & \textbf{1/8×} & \textbf{1×} & \textbf{1/5×} \\
        \bottomrule[1.2pt]
        \end{tabular}
    }
    \end{center}
\end{table}

\newpage

\section{Additional Illustration of Hallucination Evaluation}
\label{sec:appendix_demo}
\begin{figure*}[hb]
    \centering
    \begin{tabular}{@{}c@{\hspace{3mm}}c@{}}
        \includegraphics[width=0.485\linewidth]{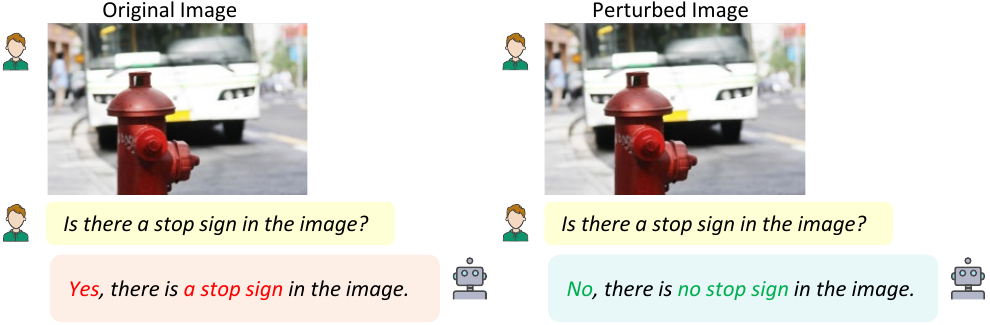}&
        \includegraphics[width=0.485\linewidth]{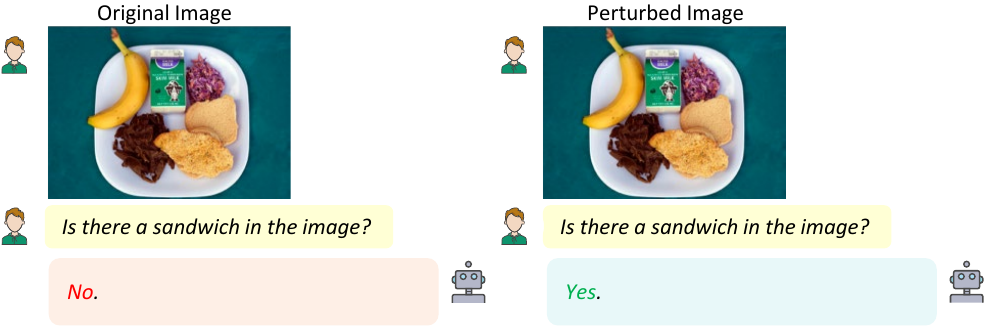}\\
        \parbox{0.485\linewidth}{\centering \footnotesize{\ \  (a) Instruct-BLIP}} & 
        \parbox{0.485\linewidth}{\centering \footnotesize{\ \  (b) LLaVA-OV}} \\
        \includegraphics[width=0.485\linewidth]{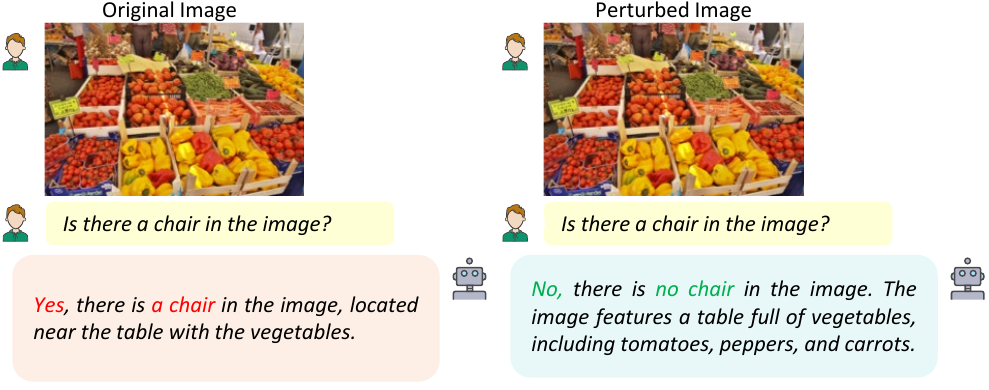}&
        \includegraphics[width=0.485\linewidth]{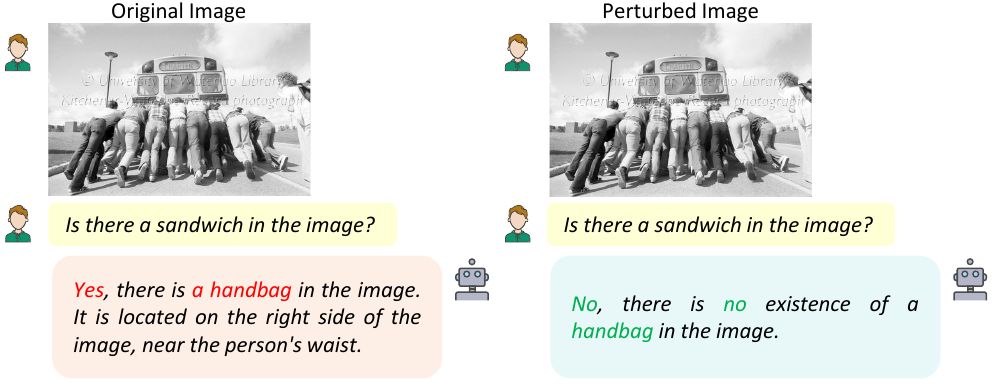}\\
        \parbox{0.485\linewidth}{\centering \footnotesize{\ \  (c) LLaVA-v1.5}} & 
        \parbox{0.485\linewidth}{\centering \footnotesize{\ \  (d) Qwen-VL2}}\\
        \includegraphics[width=0.485\linewidth]{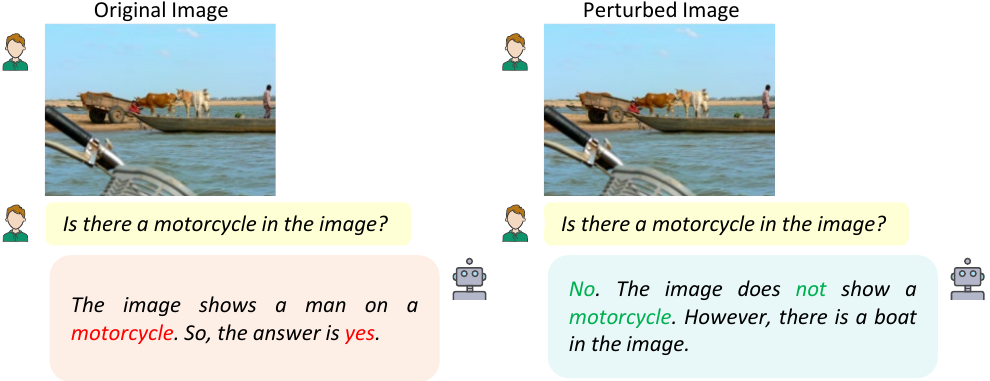} &
        \includegraphics[width=0.485\linewidth]{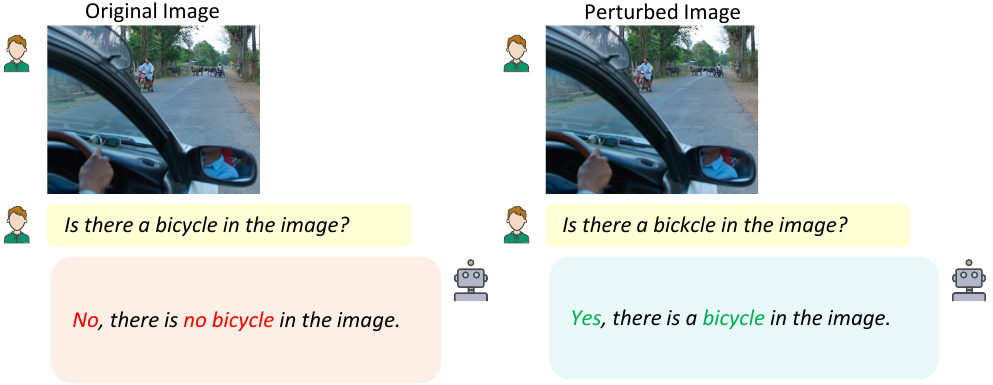}\\
        \parbox{0.485\linewidth}{\centering \footnotesize{\ \  (e) Intern-VL2}} & 
        \parbox{0.485\linewidth}{\centering \footnotesize{\ \  (f) DeepSeek-VL2}} \\
        \includegraphics[width=0.485\linewidth]{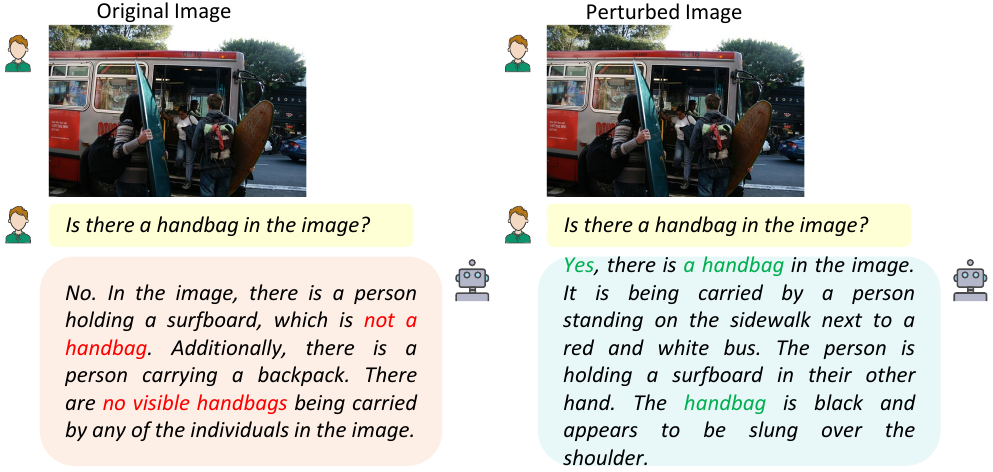} &
        \includegraphics[width=0.485\linewidth]{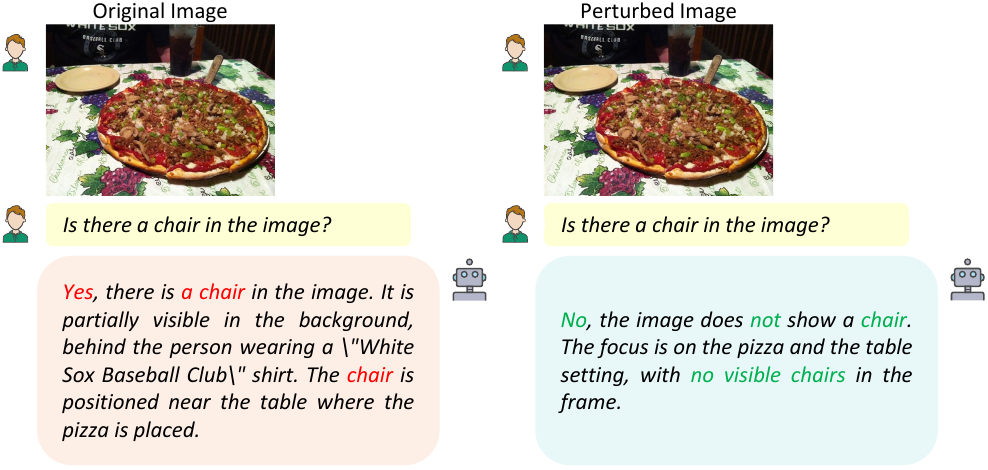}\\
        \parbox{0.485\linewidth}{\centering \footnotesize{\ \  (g) Intern-VL2-MPO}} & 
        \parbox{0.485\linewidth}{\centering \footnotesize{\ \  (h) Ovis1.6-Gemma2}}
    \end{tabular}
    \vspace{-0em}
    \caption{Illustrative examples from the POPE hallucination evaluation across eight large vision-language models: (a) Instruct-BLIP, (b) LLaVA-OV, (c) LLaVA-v1.5, (d) Qwen-VL2, (e) Intern-VL2, (f) DeepSeek-VL2, (g) Intern-VL2-MPO, and (h) Ovis1.6-Gemma2. The figure presents representative comparisons between original images and perturbed images enhanced with VAP, highlighting the differences in model responses.}
    \label{fig:appendix_demo}
    \vspace{-0.0em}
\end{figure*}

\Cref{fig:appendix_demo} presents comprehensive hallucination evaluation examples from eight state-of-the-art LVMs, demonstrating the effectiveness of our proposed method across diverse model types. While different models exhibit varying response behaviors, our approach consistently mitigates hallucinations across all cases. Notably, in models such as Intern-VL2-MPO and Ovis1.6-Gemma2, our method not only corrects erroneous responses but also facilitates the generation of more factually accurate reasoning. 
Moreover, our observations reveal that certain models exhibit fixed template-like responses to queries, such as LLaVA-OV, which provides binary responses devoid of visual context. This characteristic underscores the challenges in improving performance for such models, as their outputs of this nature pose difficulties in adversarial optimization scenarios. These results substantiate the effectiveness of the introduced visual noise VAP in alleviating hallucinations during the inference process, helping LVMs to achieve more reliable and content-aware predictions by reducing their reliance on spurious correlations and enhancing their focus on visually grounded evidence.

\newpage

\section{Orthogonality with Other Methods}
\label{sec:orthogonality}
Unlike conventional model-centric approaches, our proposed method introduces a novel paradigm for hallucination mitigation by leveraging the very mechanisms responsible for hallucinations to effectively suppress them. This innovative approach provides a fresh perspective on addressing object hallucinations in LVMs and is orthogonal to existing methods, demonstrating its potential to further alleviate hallucinations when integrated with complementary techniques. As illustrated in \Cref{st:combination_VCD}, the combination of our method with VCD~\cite{Leng_2024_CVPR} yields enhanced performance, further validating the effectiveness of our approach in mitigating model hallucinations.

\begin{table*}[h]
    \begin{center}
    \caption{Comparison of object hallucination mitigation methods under the CHAIR setting. We evaluate different techniques applied to 8 LVMs, including our proposed VAP method and its combination with visual contrastive decoding (VCD)~\cite{Leng_2024_CVPR}. $\boldsymbol{I_1}$ corresponds to: \textit{Generate a short caption of the image}, while $\boldsymbol{I_2}$ corresponds to: \textit{Provide a brief description of the given image}. CHAIR$_I$ and CHAIR$_S$ measure object hallucinations at different levels, where lower values indicate better performance. Green highlighted values represent percentage improvements relative to the preceding method, demonstrating the effectiveness of our approach and its complementarity with existing techniques.}
    \vspace{-0em}
    \label{st:combination_VCD}
    \resizebox{0.6\linewidth}{!}
    {
        \begin{tabular}{lcllll}
        \toprule[1.2pt] 
        \multicolumn{1}{l}{\multirow{2}{*}{\textbf{LVM}}} & \multicolumn{1}{c}{\multirow{2}{*}{\textbf{Method}}} & \multicolumn{2}{c}{$\boldsymbol{I_1}$} & \multicolumn{2}{c}{$\boldsymbol{I_2}$}  \\
        \cmidrule(lr){3-4} \cmidrule(lr){5-6}
         & &\textbf{ CHAIR$_I$} $\downarrow$ &\textbf{ CHAIR$_S$} $\downarrow$ & \textbf{CHAIR$_I$} $\downarrow$ & \textbf{CHAIR$_S$} $\downarrow$  \\ 
        \midrule[0.8pt]
        \multicolumn{1}{l}{\multirow{3}{*}{LLaVA-v1.5}}
        & \textit{Regular} &3.97&6.60&4.01&6.90 \\
        & \textit{VAP}   &{3.82}\G{-0.15}&{6.50}\G{-0.10}&{3.86}\G{-0.15}&{6.50}\G{-0.40} \\
        & \textit{VAP+VCD}   &\GG{3.55}{-0.27}&\GG{6.20}{-0.30}&\GG{3.67}{-0.19}&\GG{6.30}{-0.20} \\\midrule[0.4pt]
        \multicolumn{1}{l}{\multirow{3}{*}{Instruct-BLIP}} 
        & \textit{Regular} &1.83&2.90&2.14&3.40 \\
        & \textit{VAP}   &{1.71}\G{-0.12}&{2.70}\G{-0.20}&{1.96}\G{-0.18}&{3.10}\G{-0.30} \\ 
        & \textit{VAP+VCD}   &\GG{1.60}{-0.11}&\GG{2.50}{-0.20}&\GG{1.89}{-0.07}&\GG{2.70}{-0.40}
        \\ \midrule[0.4pt]
        \multicolumn{1}{l}{\multirow{3}{*}{Intern-VL2}} 
        & \textit{Regular} &4.90&7.50&5.14&9.50 \\
        & \textit{VAP}   &{4.22}\G{-0.68}&{6.60}\G{-0.90}&{4.65}\G{-0.49}&{8.90}\G{-0.60} \\ 
        & \textit{VAP+VCD}   &\GG{3.96}{-0.26}&\GG{6.10}{-0.50}&\GG{4.37}{-0.28}&\GG{8.50}{-0.40} 
        \\ \midrule[0.4pt]
        \multicolumn{1}{l}{\multirow{3}{*}{Intern-VL2-MPO}} 
        & \textit{Regular} &5.53&8.90&6.35&13.40 \\
        & \textit{VAP}   &{5.39}\G{-0.14}&{8.60}\G{-0.30}&{6.17}\G{-0.18}&{12.60}\G{-0.80} \\ 
        & \textit{VAP+VCD}   &\GG{5.14}{-0.25}&\GG{8.40}{-0.20}&\GG{6.07}{-0.10}&\GG{12.00}{-0.60} \\
        \midrule[0.4pt]
        \multicolumn{1}{l}{\multirow{3}{*}{DeepSeek-VL2}} 
        & \textit{Regular} &2.00&2.60&1.84&4.50 \\
        & \textit{VAP}   &{1.94}\G{-0.06}&{2.20}\G{-0.40}&{1.66}\G{-0.18}&{4.30}\G{-0.20} \\ 
        & \textit{VAP+VCD}   &\GG{1.89}{-0.05}&\GG{2.20}{-0.20}&\GG{1.60}{-0.06}&\GG{4.20}{-0.10} \\ \midrule[0.4pt]
        \multicolumn{1}{l}{\multirow{3}{*}{Qwen-VL2}} 
        & \textit{Regular} &3.27&5.20&3.45&6.20 \\
        & \textit{VAP}   &{2.98}\G{-0.29}&{4.80}\G{-0.40}&{3.23}\G{-0.22}&{5.70}\G{-0.50} \\ 
        & \textit{VAP+VCD}   &\GG{2.75}{-0.24}&\GG{4.50}{-0.30}&\GG{3.09}{-0.14}&\GG{5.50}{-0.20} \\ \midrule[0.4pt]
        \multicolumn{1}{l}{\multirow{3}{*}{LLaVA-OV}} 
        & \textit{Regular} &1.96&3.30&2.71&4.50 \\
        & \textit{VAP}   &{1.85}\G{-0.11}&{3.10}\G{-0.20}&{2.41}\G{-0.30}&{4.20}\G{-0.30} \\ 
        & \textit{VAP+VCD}   &\GG{1.80}{-0.05}&\GG{3.00}{-0.10}&\GG{2.33}{-0.08}&\GG{4.00}{-0.20} \\ \midrule[0.4pt]
        \multicolumn{1}{l}{\multirow{3}{*}{Ovis1.6-Gemma2}} 
        & \textit{Regular} &4.07&6.30&5.80&14.50 \\
        & \textit{VAP}   &{3.90}\G{-0.17}&{6.20}\G{-0.10}&{5.56}\G{-0.24}&{14.30}\G{-0.20} \\
        & \textit{VAP+VCD}   &\GG{3.78}{-0.12}&\GG{6.00}{-0.20}&\GG{5.39}{-0.17}&\GG{14.00}{-0.30} \\
        \bottomrule[1.2pt]
        \end{tabular}
    }
    \end{center}
\end{table*}

\section{Algorithm Details of VAP}
\label{sec:appendix_alg}
\Cref{alg:VAP} provides the detailed procedure for our proposed visual adversarial perturbation (VAP) method. To mitigate object hallucinations in LVMs, VAP optimizes adversarial perturbations by aligning the model's responses more closely with the visual content while reducing the influence of parametric knowledge bias. Given the autoregressive nature of LVMs, we employ a zero-gradient estimation strategy to optimize the perturbation direction. Specifically, our method samples perturbations over $N$ queries and leverages zeroth-order optimization to approximate the gradient of the adversarial loss with respect to the original image, enabling effective perturbation estimation in a fully black-box setting. This ensures that our approach does not require modifications to the internal inference procedure of complex LVMs. Finally, the computed perturbation is projected onto a bounded constraint $\mathbb{B}(\epsilon)$ before being applied to the input, generating a perturbed image that better satisfies the adversarial loss objectives, thereby effectively mitigating object hallucinations.

\begin{algorithm}[htbp]
    \caption{\textit{Visual Adversarial Perturbation (VAP)}}
    \label{alg:VAP}
    \begin{algorithmic}[1]
        \newcommand{\KNOWLEDGE}[1]{\item[]\hspace*{-1.55em}\textbf{Adversarial Knowledge:} #1}
        \newcommand{\PARAM}[1]{\item[]\hspace*{-1.55em}\textbf{Adversarial Parameter Setting:} #1}
        \newcommand{\GRADIENT}[1]{\item[]\hspace*{-1.55em}\textbf{Zero-Gradient Setting:} #1}
        \KNOWLEDGE{Original image $x$, Query prompt $c$, LVM $f_\theta$, Null text $\emptyset$, CLIP Text encoder $g_\psi$.}
        \PARAM{Noise magnitude $\epsilon$, Distorted timestep $T$, Noise scheduling $\mu$, step size $\alpha$.}
        \GRADIENT{Number of queries $N$, Sampling variance $\beta$, Sampling noise $\gamma$.}
        \STATE Generate a distorted image:
        \begin{equation}
            \bar{x} \sim \mathcal{N}(\sqrt{\mu_T}x, (1-\mu_T)\mathbf{I}).
        \end{equation}
        \STATE Compute initial responses:
        \begin{align}
            r_1^{(0)} = f_\theta(x, c), \\
            r_2^{(0)} = f_\theta(x, \emptyset), \\
            r_3 = f_\theta(\bar{x}, \emptyset).
        \end{align}
        \STATE Compute initial adversarial loss:
            \begin{align}
                \mathcal{L}_{s_1}^{(0)} &= \max g_\psi(r_1^{(0)})^\top g_\psi(r_2^{(0)}), \\
                \mathcal{L}_{s_2}^{(0)} &= \min g_\psi(r_1^{(0)})^\top g_\psi(r_3), \\
                \mathcal{L}_{s_3}^{(0)} &= \min g_\psi(r_2^{(0)})^\top g_\psi(r_3).
            \end{align}
        \STATE Compute overall initial loss:
            \begin{equation}
                \mathcal{L}_S^{(0)} = \frac{\mathcal{L}_{s_1}^{(0)}}{\sigma_1^2} + \frac{\mathcal{L}_{s_2}^{(0)}}{\sigma_2^2} + \frac{\mathcal{L}_{s_3}^{(0)}}{\sigma_3^2}.
            \end{equation}
        \FOR{each zero-gradient optimization step $n \in \{1, \dots, N\}$}
            \STATE Sample perturbation: 
            \begin{equation}
                \gamma_n \sim P(\gamma), \ \text{s.t.} \ \mathbb{E}[\gamma^\top \gamma] = I.
            \end{equation}
            \STATE Compute perturbed responses: 
            \begin{align}
                r_1^{(n)} = f_\theta(x + \beta\cdot \gamma_n, c), \\
                r_2^{(n)} = f_\theta(x + \beta\cdot \gamma_n, \emptyset).
            \end{align}
            \STATE Compute adversarial losses:
            \begin{align}
                \mathcal{L}_{s_1}^{(n)} &= \max g_\psi(r_1^{(n)})^\top g_\psi(r_2^{(n)}), \\
                \mathcal{L}_{s_2}^{(n)} &= \min g_\psi(r_1^{(n)})^\top g_\psi(r_3), \\
                \mathcal{L}_{s_3}^{(n)} &= \min g_\psi(r_2^{(n)})^\top g_\psi(r_3).
            \end{align}
            \STATE Compute overall adversarial loss:
            \begin{equation}
                \mathcal{L}_S^{(n)} = \frac{\mathcal{L}_{s_1}^{(n)}}{\sigma_1^2} + \frac{\mathcal{L}_{s_2}^{(n)}}{\sigma_2^2} + \frac{\mathcal{L}_{s_3}^{(n)}}{\sigma_3^2}.
            \end{equation}
            \ENDFOR
        \STATE Estimate perturbation direction via zeroth-order optimization:
        \begin{equation}
            \delta = \frac{1}{N \cdot \beta} \sum_{n=1}^N\{\mathcal{L}_S^{(n)} - \mathcal{L}_S^{(0)} \} .
        \end{equation}
        \STATE Project perturbation onto $\delta \leftarrow \text{Proj}_{\mathbb{B}_\epsilon(x)}(\delta).$
        \STATE \textbf{Return response under VAP:} 
        \begin{equation}
            w_{(VAP)} = f_\theta(\hat{x},c) = f_\theta(x+\alpha \cdot \delta,c).
        \end{equation}
        \end{algorithmic}
    \end{algorithm}

\section{Discussion}

\subsection{Analysis of False Drop Samples}

In our efforts to mitigate hallucinations in LVMs, we observed an important trade-off: the introduction of VAP occasionally leads to false drops in instances where the model's initially correct responses become incorrect after applying VAP. To quantify this phenomenon, we define two key metrics:

\begin{itemize}
    \item \textbf{False Drop Rate}: The percentage of samples where the model's initially correct responses become incorrect after applying VAP.
    \item \textbf{Correction Rate}: The percentage of samples where the model's initially incorrect responses are corrected after applying VAP.
\end{itemize}

\begin{table*}[t]
    \begin{center}
    \caption{Analysis of false drop samples and correction rates across eight LVMs on the POPE evaluation setting~\cite{pope}, using 1,000 randomly sampled MS-COCO images. False Drop Rate indicates the percentage of originally correct answers that become incorrect after applying VAP, while Correction Rate shows the percentage of originally incorrect answers that become correct. Yes Ratio Change demonstrates the shift in yes response rates before and after applying VAP in false drop samples.}
    \vspace{-0em}
    \label{st:discussion}
    \resizebox{0.55\linewidth}{!}
    {
        \begin{tabular}{lccc}
        \toprule[1.2pt] 
        \textbf{LVM} & \textbf{False Drop Rate} & \textbf{Correction Rate} & \textbf{Yes Ratio Change} \\ \midrule[0.8pt]
        LLaVA-v1.5 &0.7\%&2.1\%&85.3\% $\to$14.7\%\\ \midrule[0.4pt]
        Instruct-BLIP&0.6\%&1.3\%& 53.8\% $\to$ 46.2\% \\ \midrule[0.4pt]
        Intern-VL2&1.7\%&3.9\%&72.6\% $\to$ 27.4\% \\ \midrule[0.4pt]
        Intern-VL2-MPO&1.3\%&3.0\%&54.9\% $\to$ 45.1\% \\ \midrule[0.4pt]
        DeepSeek-VL2&0.3\%&0.9\%&57.9\% $\to$ 42.1\% \\ \midrule[0.4pt]
        Qwen-VL2&0.8\%&1.7\%&55.6\% $\to$ 44.4\% \\ \midrule[0.4pt]
        LLaVA-OV&0.5\%&1.1\%&73.3\% $\to$ 26.7\% \\ \midrule[0.4pt]
        Ovis1.6-Gemma2&0.3\%&1.1\%& 66.7\% $\to$ 33.3\% \\
        \bottomrule[1.2pt]
        \end{tabular}
    }
    \end{center}
\end{table*}

Our analysis reveals that this phenomenon is intricately linked to LVMs' parametric knowledge bias. As shown in Table~\ref{st:discussion}, we conducted comprehensive experiments across eight state-of-the-art LVMs using 1,000 randomly sampled MS-COCO images under the POPE setting. The results reveal several important insights:

First, the false drop rates remain consistently low across all models (0.3\%-1.7\%), while the correction rates are consistently higher (0.9\%-3.9\%). This favorable ratio suggests that our method's benefits substantially outweigh its potential drawbacks. Notably, newer architectures like DeepSeek-VL2 and Ovis1.6-Gemma2 achieve the lowest false drop rates (0.3\%), demonstrating the compatibility of our approach with advanced model designs.

Second, we observe a shift in the models' response patterns. The ``Yes Ratio Change'' column in \Cref{st:discussion} reveals a substantial reduction in affirmative responses. For instance, LLaVA-v1.5's "yes" responses decreased from 85.3\% to 14.7\%. This shift suggests that VAP effectively reduces the reliance on language priors, encouraging more vision-grounded responses.

Importantly, our detailed analysis reveals a critical insight into the nature of false drop cases. Recent studies have shown that LVMs exhibit a strong bias toward affirmative responses, often generating ``Yes'' responses without genuinely referring to the given vision input~\cite{ye2024beaf}. This suggests that many initially ``correct'' responses may represent lucky guesses driven by this inherent bias rather than true visual understanding. Our Yes Ratio Change statistics in Table~\ref{st:discussion} provide strong evidence for this phenomenon that the dramatic reduction in affirmative responses across all models (e.g., from 85.3\% to 14.7\% in LLaVA-v1.5) indicates that VAP effectively mitigates this bias.

This interpretation is further validated by our BEAF experimental results (Table~\ref{st:SOTA_BEAF}), where we observe significant improvements in True-Understanding (TU) metrics after applying VAP. The enhanced TU scores demonstrate that our method successfully redirects the model's attention toward question-relevant image regions, fostering genuine visual comprehension rather than reliance on statistical patterns in the training data. While this shift occasionally results in false drops, we argue that these cases represent a necessary trade-off in the transition from superficial pattern matching to visual reasoning. The consistent improvement in TU metrics across different models suggests that VAP successfully pushes LVMs toward more vision-grounded decision-making, even if it occasionally disrupts previously ``correct'' but potentially unreliable responses.

\subsection{Understanding the Effectiveness of VAP}

The consistent performance improvements across different LVMs and evaluation frameworks raise an important question: why does VAP effectively mitigate hallucinations? Our analysis reveals key mechanisms underlying VAP's effectiveness:

\paragraph{Balancing Visual and Language Signals} The success of VAP can be primarily attributed to its ability to rebalance the interaction between visual and language processing in LVMs. This is evidenced by both the significant reduction in affirmative responses (\Cref{st:discussion}) and performance improvements in vision-/text-axis hallucination assessments (\Cref{st:SOTA_BEAF}). The BEAF evaluation framework particularly demonstrates how VAP effectively interrupts the model's default reliance on parametric knowledge. The carefully calibrated perturbations strengthen visual signals during the inference process, compelling the model to ground its responses more firmly in visual evidence rather than language priors.

\paragraph{Adaptive Adversarial Noise Generation} The effectiveness of VAP is further enhanced by its adaptive noise generation mechanism. Unlike traditional adversarial perturbations that aim to maximally disrupt model predictions, VAP generates ``beneficial noise'' through zero-gradient optimization that aligns response with grounding vision input and mitigates parametric knowledge bias. This selective enhancement is validated across multiple evaluation dimensions: (1) Closed VQA format evaluations through both text-axis (POPE) and vision-/text-axis (BEAF) settings, and (2) Open-ended task evaluation through image caption generation (CHAIR). The consistent improvements across these diverse evaluation settings demonstrate VAP's ability to enhance visual understanding while maintaining task performance.

\paragraph{Architecture-Agnostic Enhancement} Our experiments across different model architectures reveal that VAP's effectiveness is not tied to specific architectural choices. This architecture-agnostic nature can be explained by VAP's operation at the input level: it modifies the visual input distribution to better align with the model's learned visual-semantic mappings, regardless of the specific implementation details. This explanation is supported by the consistent performance improvements observed across models with varying architectures, ranging from pure transformer-based models to hybrid architectures across all three evaluation frameworks (POPE, BEAF, and CHAIR).

The combination of these mechanisms creates a powerful technique for hallucination mitigation:
\begin{itemize}
    \item The rebalancing of visual-language interaction enhances visual perception while reducing spurious correlations stemming from biased language priors.
    \item The adaptive adversarial visual noise generation employs strategic optimization to influence LVM decision processes, ensuring that perturbations enhance rather than compromise visual understanding.
    \item VAP operates in a completely black-box manner requiring no access or modification to the LVM, establishing it as a broadly applicable solution across different model architectures.
\end{itemize}

%%%%%%%%%%%%%%%%%%%%%%%%%%%%%%%%%%%%%%%%%%%%%%%%%%%%%%%%%%%%%%%%%%%%%%%%%%%%%%%
%%%%%%%%%%%%%%%%%%%%%%%%%%%%%%%%%%%%%%%%%%%%%%%%%%%%%%%%%%%%%%%%%%%%%%%%%%%%%%%

\end{document}